\documentclass[10pt,journal,compsoc]{IEEEtran}

\usepackage{tabulary,graphicx,amsmath,amsfonts,amssymb, adjustbox}
\usepackage{subfigure}
\usepackage{ragged2e}
\usepackage{multirow}
\usepackage{dblfloatfix}
\usepackage{blindtext}
\usepackage{booktabs,caption}
\usepackage{float}
\usepackage[utf8x]{inputenc}
\usepackage{amsmath}
\usepackage{mathtools}

\ifCLASSOPTIONcompsoc
  \usepackage[nocompress]{cite}
\else
  \usepackage{cite}
\fi
\ifCLASSINFOpdf
\else
\fi
\newcommand{\cmmnt}[1]{}
\begin{document}
%
\title{Deep Learning Analysis and Age Prediction from Shoeprints}
\newcommand{\negpar}[1][-1em]{%
	\ifvmode\else\par\fi
	{\parindent=#1\leavevmode}\ignorespaces
}
%
%
%
%

\author{Muhammad~Hassan, Yan~Wang, Di~Wang, Daixi~Li, 
	Yanchun~Liang, You~Zhou$^{\textbf{*}}$, and Dong~Xu$^{\textbf{*}}$
	\IEEEcompsocitemizethanks{\IEEEcompsocthanksitem M. Hassan, Y. Wang, Y. Liang, and Y. Zhou are with the College of Computer Science and Technology, Jilin University, Changchun, China.\protect
	E-mail: mhassandev@gmail.com, wy6868@jlu.edu.cn,ycliang@jlu.edu.cn, zyou@jlu.edu.cn
	\IEEEcompsocthanksitem D. Wang is a member of Joint NTU-UBC Research Centre of Excellence in Active Living for the Elderly, Nanyang Technological University, Singapor.\protect
	E-mail: WangDi@ntu.edu.sg
	\IEEEcompsocthanksitem D. Li is the CEO of Everspray Science and Technology Company Ltd., Dalian, China.\protect
	E-mail: lidaixi@everspry.com
	\IEEEcompsocthanksitem D. Xu is a member of AAAS, AIMBE, Department of Electrical Engineering and Computer Science, University of Missouri-Columbia
	Columbia, USA.\protect
	E-mail: xudong@missouri.edu
}
	\thanks{}}

%
%

\markboth{}%
{Shell \MakeLowercase{\textit{et al.}}: Bare Advanced Demo of IEEEtran.cls for IEEE Computer Society Journals}
%



\IEEEtitleabstractindextext{%
\begin{abstract}
Human walking and gaits involve several complex body parts and are influenced by personality, mood, social and cultural traits, and aging. These factors are reflected in shoeprints, which in turn can be used to predict age, a problem not systematically addressed using any computational approach. We collected 100,000 shoeprints of subjects ranging from 7 to 80 years old and used the data to develop a deep learning end-to-end model ShoeNet to analyze age-related patterns and predict age. The model integrates various convolutional neural network models together using a skip mechanism to extract age-related features, especially in pressure and abrasion regions from pair-wise shoeprints. The results show that 40.23\% of the subjects had prediction errors within 5-years of age and the prediction accuracy for gender classification reached 86.07\%. Interestingly, the age-related features mostly reside in the asymmetric differences between left and right shoeprints. The analysis also reveals interesting age-related and gender-related patterns in the pressure distributions on shoeprints; in particular, the pressure forces spread from the middle of the toe toward outside regions over age with gender-specific variations on heel regions. Such statistics provide insight into new methods for forensic investigations, medical studies of gait-pattern disorders, biometrics, and sport studies.
\end{abstract}

\begin{IEEEkeywords}
Shoeprint, Gait-and-standing patterns, Aging, Age prediction, Deep learning, Pressure distribution.
\end{IEEEkeywords}}

\maketitle

\IEEEdisplaynontitleabstractindextext

%
\IEEEpeerreviewmaketitle
\IEEEraisesectionheading{\section{Introduction}\label{sec:introduction}}

\IEEEPARstart{A}{}shoeprint is the impression mark made by the contact of footwear tread with the ground surface\cite{49}. Such impressions in the digitized format can be captured from pressed regions of footwear in a standing or walking position\cite{52,104}. Retrieving shoeprints is a challenging computer vision problem given various factors such as textures, designs, manufacture models and wear-effects\cite{105}. Many processing methods have been applied for retrieving shoeprints, including manual\cite{10}, semi-automated\cite{6,23}, and automated\cite{107,36} approaches, especially machine-learning\cite{69,21,32} based automated methods. Recently, deep learning methods have been applied in processing, which include features extraction, reconstruction and investigation of shoeprints; furthermore, these methods have achieved some good results\cite{69,21,32,123} leading to more extensive studies on shoeprints matching\cite{29}, recognition\cite{55}, and reconstruction\cite{21,75}. Some of these studies are carried out for crime-scene investigation, such as forensic podiatry\cite{29,51,58,69,106} while others focus on biological trait examination and investigation, such as stature estimation\cite{40}, gender prediction\cite{41,8,106} and body morphology examination\cite{67}. Unlike other biometric modalities, such as fingerprints, palm-prints, and retina prints, shoeprints have inconsistent patterns, shape, and appearance, except for wear-effects. Such effects are associated with distinct traits including gait patterns, personality, mood, social and cultural variances, age, and gender\cite{17,25,37,38} which provides a basis for crime-scene investigations, perpetrator gait analyses, and sport and health examinations.

Among biological profiles, age is a common trait that can be reflected in shoeprints. In contrast to age predictions based on other human traits including facial images\cite{91,108} brain MRI and EEG\cite{92,93, 95}, DNA\cite{94} and gait patterns\cite{2,35}, age prediction from shoeprints is more challenging due to image noises, varying patterns, and shoe designs. Most importantly, wear-effects on shoeprints are influenced by various biomechanics such as the weight, gait patterns, environmental factors, personality, and tread materials of the shoes\cite{21}. Hence, although forensic experts often estimate the age of the shoe owners  from shoeprints empirically, no systematic computational method has been developed to predict age based on shoeprints. To predict age based on shoeprints systematically, it is essential to study their relationships as shown by age, gaits, standing patterns, and shoeprints. Aging is significantly associated with gait patterns in the capacity of stances, bipedal floor contact, step length, posture stooping during walking, gender and ethnicity\cite{3,5,35,44,57}. It is known that gait speed\cite{53}, step rate\cite{34}, stance and gait-based widening\cite{35} are related to aging\cite{40,89}. Some studies have reported that in normal walking, elder people (age 
$>$ 65 years) prefer 41\% wider steps than younger subjects (age $<$ 30 years)\cite{2,38}. Furthermore, such variational effects are also reflected in the pressure distributions of the footwear as muscles recalibrate themselves over age. The energy cost for muscle movements during walking is high in children and steadily decreases over age\cite{61,100} which can be reflected in the imprinted shoeprints of healthy subjects\cite{2,12,44}. So, the wear-effects with natural erosion on shoeprints have the potential to be used for predicting age by involving distinct contact regions of the footwear\cite{12,31,45,65}.
\begin{figure*}[!]
	\centering
	\includegraphics[height=2.8in,width=6.2in]{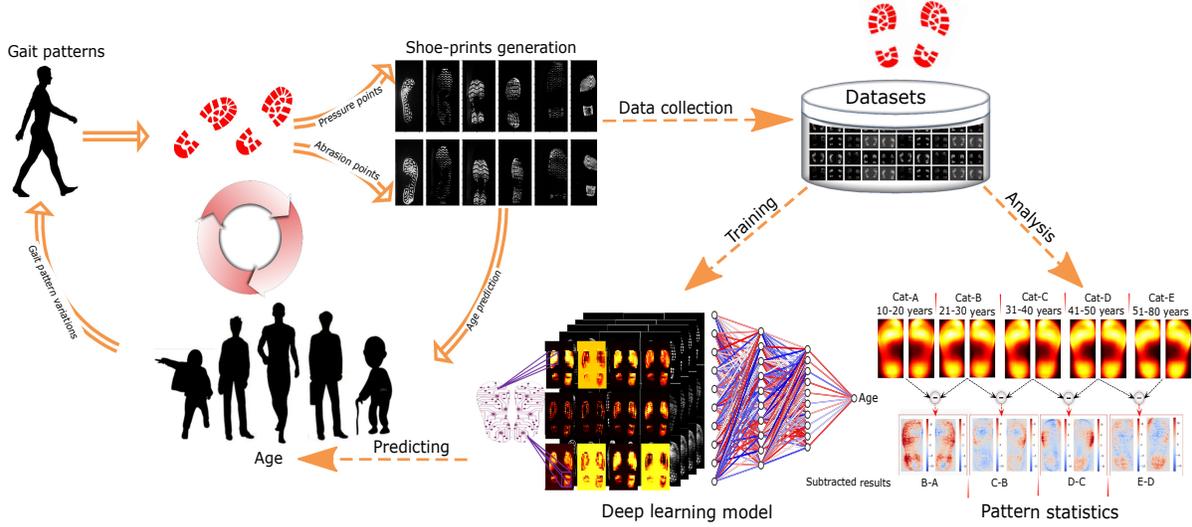}
	\caption{\normalfont{The workflow of this study for age prediction and analysis from shoeprints. The relationship (O-shape) among aging, gait/standing patterns, and shoeprints. Based on the data collection for training deep learning models, the age prediction model is trained. The core model is ShoeNet to predict age. Variations of pressure forces on shoeprints versus age progression as well as the plots of region-wise trends are obtained from the data.}}
	\label{fig:triangularity}
\end{figure*}

In this study, we explored and analyzed the relationship between aging and shoeprints and used the relationship to predict age from shoeprints using a deep learning approach, as shown in Figure ~\ref{fig:triangularity}. Our contributions to this new research area include: 1) To the best of our knowledge, this study represents the first effort to systematically explore the relationship among aging, gait pattern, and shoeprints. 2) We collected 100,000 images featuring 50,000 subjects between the ages of 7 and 80 years old thereby providing a large-scale annotated shoeprint dataset for the research community to use in related studies. 3) We proposed a convolution neural network (CNN) adapted model named ShoeNet based on the comparative studies of different deep learning approaches to predict age, and 4) we statistically analyzed group-wise pressure distributions based on age. Our findings successfully revealed the association of biological profiles with the wear-effects reflected on shoeprints, which can be used for biological profile estimations, forensic investigations, and examining sports and health activities.
\section{Methodology}
\subsection{Datasets acquisition and generation.} We collected shoeprints by having subjects step on the strip of the EverOS V2.0 (Supplementary Figure S1\cmmnt{~\ref{fig:Foot_retrival_machine}}) acquisition system. It is an intelligent footprint terminal system which collects footprint and shoeprint samples using optical-array sensors.The footprint terminal system also generates corresponding standard shoeprint pattern images in digitalized form. . The hardware structure is 402 mm long $\times$ 315 mm wide  $\times$ 152 mm-high with a collection scope of 350 mm  $\times$ 150 mm, having a USB-3.0 interface and output images in 300-DPI (Figure S1\cmmnt{ ~\ref{fig:Foot_retrival_machine}}). The equipment can preprocess the acquired image by removing the background noise. Prior to training a deep network model, the acquired images were pre-processed by resizing each of them into the same dimensions (224 H$\times$112 W), converting eacg image into a hue-saturation-value (HSV) format and masking the strip using a threshold value. To train the ShoeNet model, we flipped the images to the natural orientation and discarded poor-quality images manually. In most cases, a network model requires a balance dataset; hence, the number of samples per age range varies significantly so that the augmentation was performed to balance the sample distribution and increase the diversity of images per age range by providing an extensive amount of refined data for training deep-learning models. The augmentation operations included Gaussian-noise, flipping (left-to-right), rotation, and cropping shoeprints (Supplementary Figure S1)\cmmnt{~\ref{fig:Unprocessed_dataset})}.

There are 100,000 images in total obtained from 50,000 subjects ranging from age 7-to- 80. Each subject has two (left, right) shoeprint images. The dataset contains labeled information of age, gender, height and weight as well as rulers embedded in the side of each shoeprint. The main features of the dataset versions are listed in Supplementary Table S1\cmmnt{~\ref{tab:datasets}} and are described as follows: Dataset-A contains the original images with two (left \& right) shoeprints for each subject (Supplementary Figure S2).\cmmnt{ ~\ref{fig:Unprocessed_dataset}}. For Dataset-B, the ruler/scale attached to every image was discarded. To further refine, all the shoeprints were aligned manually in natural order as pairwise-shoeprints worn by humans. Poor quality images were discarded. Dataset-C solely holds the left shoeprints only, while Dataset-D conversely holds the right shoeprints only. Dataset-E integrates left\&right shoeprints into a single pair-wise shoeprint (Supplementary S2\cmmnt{Figure ~\ref{fig:Unprocessed_dataset}}). Dataset-F is comprised of shoeprints separated for males and females to train models for gender classification and age estimation. To classify gender, we need a balanced dataset; however, the existing dataset is imbalanced in the number of male-and-female samples. For this purpose, we performed augmentation by the inclusion of Gaussian-noise, rotating and cropping to balance the number of samples between males and females. The overall statistics, including the original dataset, before and after augmentation, training, validation and testing dataset splits are given in Supplementary Table S1\cmmnt{~\ref{tab:datasets}}. For Dataset-G, we performed the augmentation to balance the sample distribution per age range (Figure S2\cmmnt{~\ref{fig:Unprocessed_dataset}}).
\subsection{Segmentation and superimposition.}The acquired shoeprints were processed for studying group-wise pressure distributions. For each group, the images were superimposed to analyze the pressure distributions. First, a threshold value was set up after converting the color images to gray-scale images using Open CV\cite{127}. To find the bounding box of shoeprints, the images were processed for finding contours by border following algorithm. The corresponding bounding boxes were computed for each contour, pruned down contours by figuring out the minimum top-left and maximum bottom-right coordinates. The images were cropped based on the bounding-box coordinates and resized into the same aspect-ratio by the inter-cubic interpolation. All the segmented images were then superimposed into the same coordinates for all the divided categories. For performing subtraction operations, the data type of superimposed images was converted into integer-16 (i.e., -32,768 to +32,767) to keep both negative and positive results.

\subsection{Evaluation metrics.} For a regression problem of age prediction, we apply cumulative score (CS) and mean cumulative score (MCS) as evaluation metrics to accommodate the nature of the problem. CS and MCS imitate the existing studies\cite{124,125,126}, and are used to assess accuracies in a range of age groups. CS (or $CS_j$) and MCS (or $MCS−J$) give more weight to the smaller ranges of match windows. The ranges depend on the value of $j$ and $J$, the absolute differences between actual and estimated age scores, as shown in the following:
\begin{equation}
	MCS-J=\frac {\sum_{j=0}^{J} CS_j}{J+1}\label{equ:1}
\end{equation}
\begin{equation*}
	CS_j=
	\frac {\sum_{i=1}^{n} f(\delta_i)}{n}*100
\end{equation*}
	where
\begin{equation*}
	f(\delta_i)=
	\begin{cases}
	1,& \text{if	}  \delta_i\leq j\\
	0,& \text{if	}  \delta_i>j\\
	\end{cases}
\end{equation*}
$CS_j$ is the percentage mean of $f(\delta_i)$, where $f(\delta_i)$ is the Euclidean-distance $(|y_i-\bar{y_i}|)$ between actual ($y_i$) and predicted ($\bar{y_i}$) score, and it will be counted as 1 for $|y_i-\bar{y_i}|\leq j$. The value of $f(\delta_i)$ expressed as zero (0) implies that the distance $(|y_i-\bar{y_i}|)$ is greater than the threshold value ($j$). The 
$MCS$ score facilitates prediction in various ranges of matching thresholds rather than a single threshold. Thus MCS score gives a more comprehensive assessment for the challenging problem of shoeprint-based age prediction to cover all the values of $|y_i-\bar{y_i}|\leq j$ for the setup threshold ($j$). This also allows us to give different penalties with varying thresholds in the objective function of the deep learning model. 	
\subsection{Regression specific custom loss function (CLF).} Because age prediction is a regression problem, a single output will be expected as a result. For empirical significance, we customize the mean-square-error (MSE) into a specialized form (CLF) to optimize the hyperparameters during training. The under-studied optimizer fine-tunes the weights of convolution filters to minimize the loss value. To produce regression specific results, CLF penalizes the out-ranged values more. It minimizes the distance between the actual and predicted age in a target-oriented way. The formulation of CLF is illustrated in the following equation:		
\begin{equation}
	CLF=
	\frac {\sum_{i=1}^{n} E_i}{n};E_i=
	\begin{cases}
	d_i*\epsilon,& \text{if	}  d_i\leq J\\
	d_i^{3}+\epsilon,& \text{if	}  d_i>J
	\end{cases}\label{equ:2}
\end{equation}
CLF is the mean of difference $(E)$ for $n$ number of samples, where $n$ = $(total-samples)/(input$-$size)$. $\epsilon$ is a small value (0.0001-to-0.3) used to prevent the network from attaining zero difference and to sustain the learning process. Similarly, $d_i=||y-\bar{y}||$ is the Euclidean distance between actual-age $(y)$ and predicted age $(\bar{y})$. Furthermore, $J$ is a natural number derived from \textit{MCS-J} for predictable age ranges. In the second condition ($d_i\leq J$), the values above the $J$ will penalize more the weights based on the computed loss-value in the exponential time of power 3. The penalization influences the optimization of network weights and biases. It will direct the optimizer to tune these parameters in order to minimize the difference between actual and predicted age. The CLF values for $J$=$2$, $J$=$3$ and mean-square-error (MSE) are illustrated in Supplementary Figure S3\cmmnt{ ~\ref{fig:CLFvsMSE}}. The CLF values indicate abrupt changes for $J$=$2$ and $J$=$3$, which demonstrates a high penalty by following that the \textit{MCS-J} would be only counted in the given range of $J$. 
To verify the effectiveness of CLF, we carried out the training of our proposed model (ShoeNet) based on both CLF and a mean-square-error (MSE) loss function up to 500 epochs. The evaluation metrics (\textit{MCS-2, MCS-3}) is remarkably superior for ShoeNet trained with CLF (Supplementary Table S3\cmmnt{ ~\ref{tab:CLS-vs-MSE}}). CLF-based ShoeNet has higher evaluation scores compared to MSE-based trained network (Supplementary Table S3\cmmnt{ ~\ref{tab:CLS-vs-MSE}}). The customization of CLF better demonstrates convergence evaluation metrics. MSE finds the mean square difference of actual and predicted values and, therefore, gives more significant value to MAE rather than \textit{MCS-J} ($J=2,3..n$). Hence, CLF is not only considered the Euclidean-distance but also more frequently penalizes the adjacent values to $J$ in MCS-J. By more penalization, the practiced optimizer fine tunes the learning weights to obtain a persuasive estimation score. Adam is used as an optimizer with the L-2 regularizer to tune hyper-parameters as described in the Methods section.
\subsection{Deep-learning model for age prediction.} The architecture of our proposed ShoeNet adapts the skip-connections and dense structures from ResNet and DenseNet, respectively. ShoeNet is the integration of five convolution blocks (Blocks-A, B, C, D, E), three fully connected layers and a single output of linear regression for age prediction (Figure ~\ref{fig:suggested-modde}). All the stacked convolution layers inside the blocks are followed by pooling, batch-normalization (BN), and rectified linear units (ReLus). At the higher level, the fed images are passed through BN and ReLu functions prior to convolution layers. Each block is connected to the next block in two ways: first, the output of the earlier block feeds as the input to the next deep level block, for instance, Block-A inputs to Block-B, Block-B inputs to Block-C, and so on. Second, the output of each block, except for the last layer, skips and down-samples to all the deep block as skip-layers and merges with the corresponding block. For instance, block-A skips and down-samples to  the corresponding dimensions of deeper blocks (i.e., Block-C, Block-D, and so on) and merges along the specified axis with the output of Block-B, Block-C, and so on. Finally, all the skipped and convolutional layers merge channel-wise into a stack of representations and then densely connect with three fully connected layers leading to a single output as age prediction. The three fully connected layers contain 384 neurons each, with a single neuron for the last layer.

In order to convolve the feature-maps speedily and converge the weights and biases for achieving the target, we converted the color images into grayscale number of channel as one(1). After resizing, the dimensions of each image are height $H (224)$, width $W (224)$ pixels and the number of channels $N (1)$.  During ShoeNet training, the images feed as the integration of (Input\_size, $224, 224, 1$). At the higher level, by applying the BN function followed by ReLu, the training process accelerates, and brings smoothness and regularization in the loss function landscape as well as reducing the covariate shift. The three stacked shallow convolution layers are structured as $(X_i, F_n, F_s)$, where $X_i$ is the input from the previous layer, $F_n$ is the number of filters, and $F_s$ is the filter-size. For instance, the parameter arrangement for Block-A is ($224\times224, 32, 3\times3$), representing the input dimensions, number of filters and filter-size, respectively. Similarly, for Block-B ($112\times112, 64, 3\times3$), the input dimensions were down-sampled while doubling the filter size, and so on. The skip layers cumulate after merging with the next layer followed by BN and ReLu. Merging at Block-E produces $A\oplus B\oplus ... \oplus E$, where $\oplus$ corresponds to the concatenation operation (Figure ~\ref{fig:suggested-modde}). For matching the number of feature maps, each skip layer down-samples and up-samples after applying 1$\times$1 convolution to maintain learned features at a higher level and avoid significant performance degradation (Supplementary Figure S4\cmmnt{ ~\ref{fig:correspong_scratching}}). To avoid over-fitting, the fully connected layers drop out by 30\%, 40\% and 50\% rate, respectively. Moreover, the L-2 regularizer with a weight decay of 0.001 embeds in the second-last fully connected-layer to avoid large weights by pushing the weights toward zero (0). The Adam optimizer tunes the networks weights from the initial learning rate (LR) 0.001, $\beta_1$(0.99), $\beta_2$(0.999), epsilon(1e-08) and learning-decay-step 10,000 based on the loss value of CLF.
\subsection{ShoeNet for gender classification and analysis.} We examined patterns of pressure spreading on shoeprints that varied from males to females. First, we classify the gender by deploying a two-class based prediction. For gender classification and gender-based age analysis, we trained and validated ShoeNet structure with balanced dataset-F (Supplementary Table S2 \cmmnt{~\ref{tab:gender-dataset-distribution}}). As a classification problem, the same ShoeNet structure (Figure ~\ref{fig:suggested-modde}) was modified with minor changes in terms of the number of neurons in fully connected layers. After modification, the number of neurons were 512, 384 and 256 for the corresponding three fully connected layers, while the last layer had a two-neurons output with a softmax activation function to perform classification. We applied categorical cross entropy as the loss-function, with Adam optimizer, and Accuracy and Confusion as evaluation metrics. After training, we observed noticeable results with 89.34\% training and 86.07\% testing accuracy for gender prediction. The evidence in the corresponding confusion metric demonstrates the significant classification results (Supplementary Table S4\cmmnt{ ~\ref{tab:gender-testing-result}}).
\section{Results}
\subsection{Architecture of ShoeNet.} 
\begin{figure*}[ht!]
	\raggedleft
	\includegraphics[ width=6.3in, height=1.9in]{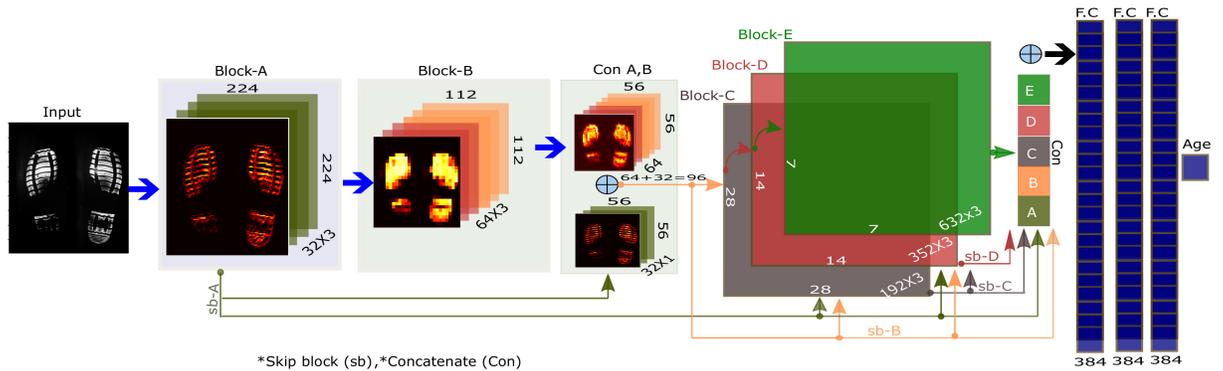}
	\caption{\textnormal{Architecture of the proposed model (ShoeNet). Each input image of shoeprints is 224 by 224 pixels. There are five blocks (Blocks A, B...E). The output of each block feeds as the input to the next block, i.e. Block-A to Block-B, and Block-B to Block-C, and so on. The internal block structure contains a stack of convolution operations; in particular, Block-A has 3 feature maps of 32 filters, and the dimension reduces by 2 fold across each block (down-sampling). Each block outputs down-samples, skips to the next block (sb) and merges (i.e. Con A,B) together to produce the input for the deep level block. The skip connections are shown with arrows, while all the blocks merge in the last layer shown with the $\oplus$ operation. After skipping and merging, there are 32 and 64 feature maps for Block-A and Block-B, respectively. While after the concatenation, the number of feature maps increases to 96. Finally, all the fully connected (FC) layers end up with a single neuron for age prediction.}}
	\label{fig:suggested-modde}
\end{figure*}

ShoeNet is designed to effectively learn the wear-effects on shoeprints varying with gait and standing patterns corresponding to age progression. It is based on pairwise shoeprints and captures extracted features from left and right shoeprints both in isolated and joint forms. ShoeNet is equipped with a skip mechanism to re-parameterize the stack of features at the next convolutional layer. ShoeNet utilizes a customized loss function in a target-oriented fashion to handle the outliers in the data. The architecture of ShoeNet is shown in Figure ~\ref{fig:suggested-modde}. ShoeNet adapts the concept of skip-connections from Residual-Net\cite{110} and a dense structure from DenseNet\cite{109}. It integrates five blocks (Blocks A, B, C, D, and E), three fully connected layers and a linear regression for age prediction (see Methods). The skip layers integrate the learned features from early levels, which help them avoid the degradation of shallow stacked networks and circumvent the gradient information loss to maintain features of gait and standing patterns in the corresponding pair-wise shoeprints (Supplementary Figure S4\cmmnt{~\ref{fig:correspong_scratching}}). After the batch normalization (BN) and the activation of ReLu, the network becomes dense via three fully connected layers of 384 neurons per-layer and finally ends with a linear operation for age prediction (Methods). ShoeNet extracts feature maps from both left and right (L\&R) shoeprints as a unified pair-wise representation to obtain the wear effects reflecting the walking characteristics of individuals. The parameter sharing in the form of skip-block enables the model to learn holistic age-related features from shoeprints.
\subsection{Data collection and preprocessing.} The shoeprints were acquired by Everspry Outsole Scanner (EverOS V2.0) as shown in Supplementary Figure S1\cmmnt{ ~\ref{fig:Foot_retrival_machine}}, over a  period of two years (2017 -to- 2018). The shoeprints were acquired by the Hengrui acquisition system (Supplementary Figure S1\cmmnt{~\ref{fig:Foot_retrival_machine}}) in two years (2017-2018). A total of 100,000 images were obtained from 50,000 subjects ranging from 7 to 80 years old. Each subject has two (left, right) shoeprints images\cite{70}. To the best of our knowledge, this is the largest scale dataset of shoeprints with annotated information of age, gender, height and weight as well as rulers embedded on one side of each shoeprint\cite{51}. A list of data generation versions was applied to facilitate a wide range of deep learning practices in terms of a biological profile estimation (Supplementary Figure S2\cmmnt{~\ref{fig:Unprocessed_dataset}}(a)). A detailed description of the data collection, preprocessing, and dataset versions can be found in the Methods section. 
\subsection{ShoeNet vs. state-of-the-art models.} 
\begin{figure*}[ht!]
	\centering
	\includegraphics[height=2.8in,width=6.5in]{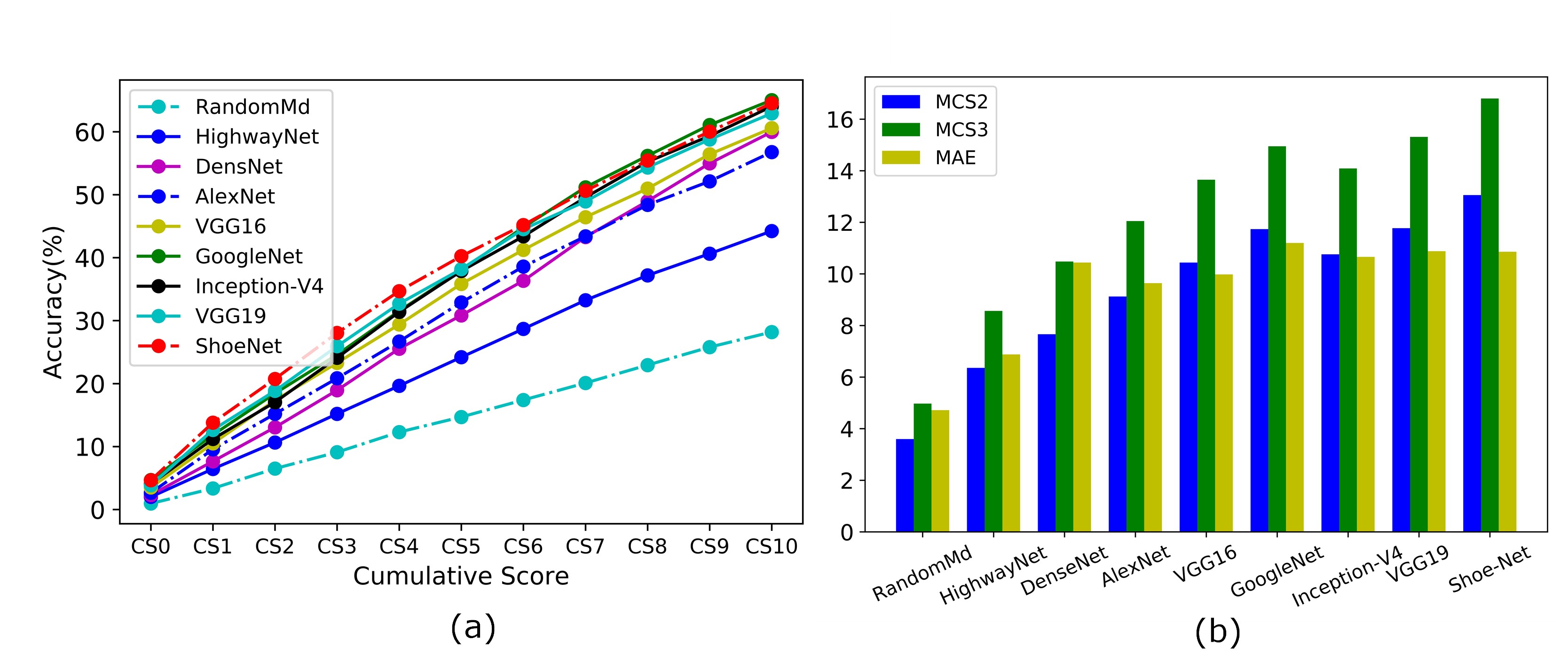}
	\caption{\normalfont{Comparison between ShoeNet and other state-of-the-art deep learning models. (a) The cumulative scores of ShoeNet and standard models are depicted in terms of CSj (for j=0,1,...10). ShoeNet has a similar trend with GoogleNet in terms of $CS_j$ (for $j>6$) but ShoeNet has a higher score of $CS_j$ (for $j\leq6$) than any other model within the prediction range. (b) The bar graph is the depiction of three evaluation parameters with standard modalities, including MAE percent-accuracy, \textit{MCS-2} and \textit{MCS-3}. The detailed statistical results are listed in Supplementary Table S5\cmmnt{~\ref{tab:Comparative-study}}}}
	\label{fig:Comparative-study-graph}
\end{figure*}

We compared the ShoeNet architecture with some state-of-the-art deep learning based models. The models for the comparative study against ShoeNet included Random-Method (RandomMd), HighwayNet\cite{112}, DenseNet\cite{109}, AlexNet\cite{113}, VGG16\cite{55}, GoogleNet\cite{115}, Inception-V4\cite{116}, and VGG19\cite{55}. To compare against a baseline, we used RandomMd, which is based on the random numbers generated to validate by-chance age prediction. All the models were trained and tested for age estimation with the same dataset and similar hyper-parameters. For performance assessment, cumulative score (CS), mean cumulative score (MCS) and mean absolute error (MAE) percent accuracy were applied (see Methods), and the comparative illustrations are shown in Figure ~\ref{fig:Comparative-study-graph}. $CS_j$  can be computed as the counting of all those subjects whose age differences (between ground-age and predicted-age) are within the given range ($for j = 0,1,2...10$), and then divided by the total number of test samples ($n$) and multiplied by 100. For instance, the true predicted score ($CS_2$) for a given range ($for j \leq 2$), will be the sum of all subjects whose age differences are within 2 years and divided by $n$, and then multiplied by 100. Similarly, \textit{MCS-J} computes the mean of $CS_0, CS_1, CS_2 … CS_J$ scores. In contrast to a single assessment value for regression, these metrics evaluate the prediction performance in a broad relevant range. Using these metrics, ShoeNet reported higher scores of \textit{MCS-2} (13.06) and \textit{MCS-3} (16.81) than other methods. ShoeNet also presented an acceptable percent accuracy score (Figure ~\ref{fig:Comparative-study-graph}(b)). The prediction of ShoeNet for the subjects in the given range $CS_5$ (within 5 years) is 40.23\%, and other accuracy scores $CS_j$  ($for j=0,1,2..10$) are shown in Figure ~\ref{fig:Comparative-study-graph}(a).
\subsection{Deep learning approaches for shoeprints based biological traits analysis.}
It is expected that pair-wise shoeprints have more age-related information in standing and gait-patterns than in an isolated single shoeprint. To better utilize the information for age prediction, a number of dataset versions were formed, and several deep learning models were developed corresponding to the nature of the datasets (Supplementary Figure S2\cmmnt{ ~\ref{fig:Unprocessed_dataset}}). The detailed descriptions of training, validation, and testing of these datasets and their modalities are placed in Supplementary Method 1. Our deep models consist of a Left\&Right shoeprint based convolution neural network (LR-CNN) (Figure ~\ref{fig:suggested-modde}), a fusion model (FM-CNN) (Supplementary Figure S5\cmmnt{ ~\ref{fig:Fusion Model}}) and a multi-model architecture (MM-CNN) (Supplementary Figure S6\cmmnt{~\ref{fig:multi-model}}). The datasets used for training LR-CNN, FM-CNN and MM-CNN are both Dataset-C and Dataset-D, while ShoeNet was trained on Dataset-G (Supplementary Table S1\cmmnt{~\ref{tab:datasets}}). LR-CNN extracts features from a single shoeprint (either left or right). FM-CNN combines dissimilar features extracted by a twin network corresponding to the two shoeprints (L\&R). Furthermore, MM-CNN incorporates parameter sharing and an ensemble approach to map the shoeprint features with human age. It contains multiple architectures and allows each architecture to benefit from isolated features. All three modalities capture complementary features. In contrast to LR-CNN, FM-CNN and MM-CNN, ShoeNet interprets both (L\&R) a single representation to estimate the age. It takes both shoeprints as a single input image to utilize the representation of gait patterns in the corresponding opposite shoeprint as illustrated in Supplementary Figure S4\cmmnt{ ~\ref{fig:correspong_scratching}}. ShoeNet outperforms all the other models with \textit{MCS-2} (13.06) and \textit{MCS-3} (16.80) while producing an acceptable result for MAE-percent-accuracy (Supplementary Table S6\cmmnt{ ~\ref{tab:tab2}}).
\subsection{Result analysis of group-wise aging.} We analyzed the significance of ShoeNet in terms of group-wise aging and consistent variations of pressure effects on shoeprints caused by gait and standing patterns. To better depict the aging association with shoeprints, we categorized shoeprints into two types, i.e. Type-A and Type-B. Type-A is divided into three sub-categories in terms of age groups, which are 10-80 (ShoeNet main-category), 20-50 and 25-45 year olds, while Type-B is sub-divided into cat-A (10-20), cat-B (21-30), cat-C (31-40), cat-D (41-50) and cat-E (50-80) year olds, as shown in Supplementary Table S7\cmmnt{~\ref{tab:tab3}}. In addition, ShoeNet associates the relevant wear-effects with respect to the group-wise age estimation. As shown in Supplementary Table S7\cmmnt{~\ref{tab:tab3}}, Type-A has the best performance in the category of 25-45 year old subjects. Similarly, in Type-B, cat-B (21-30) has highest scores in MCS-2 (17.86) and MCS-3 (22.67), which indicates that the subjects in cat-B (21-30) have more consistent effects of gait and standing-patterns on shoeprints than other age groups. Furthermore, the performance in cat-E (51-80) is relatively poor given the inconsistent gait-and-standing patterns and pressure distributions on the shoeprints. The elder subjects above 50 years have diversities in their health conditions. For instance, some people might have advanced age but healthy conditions, which may show youthful walking and gait patterns, while some other people may have severe diseases that affect walking and standing.
\subsection{Trend analysis of group-wise pressure distributions with aging.}
\begin{figure*}[h!]
	\centering
	\includegraphics[width=18cm,height=9cm]{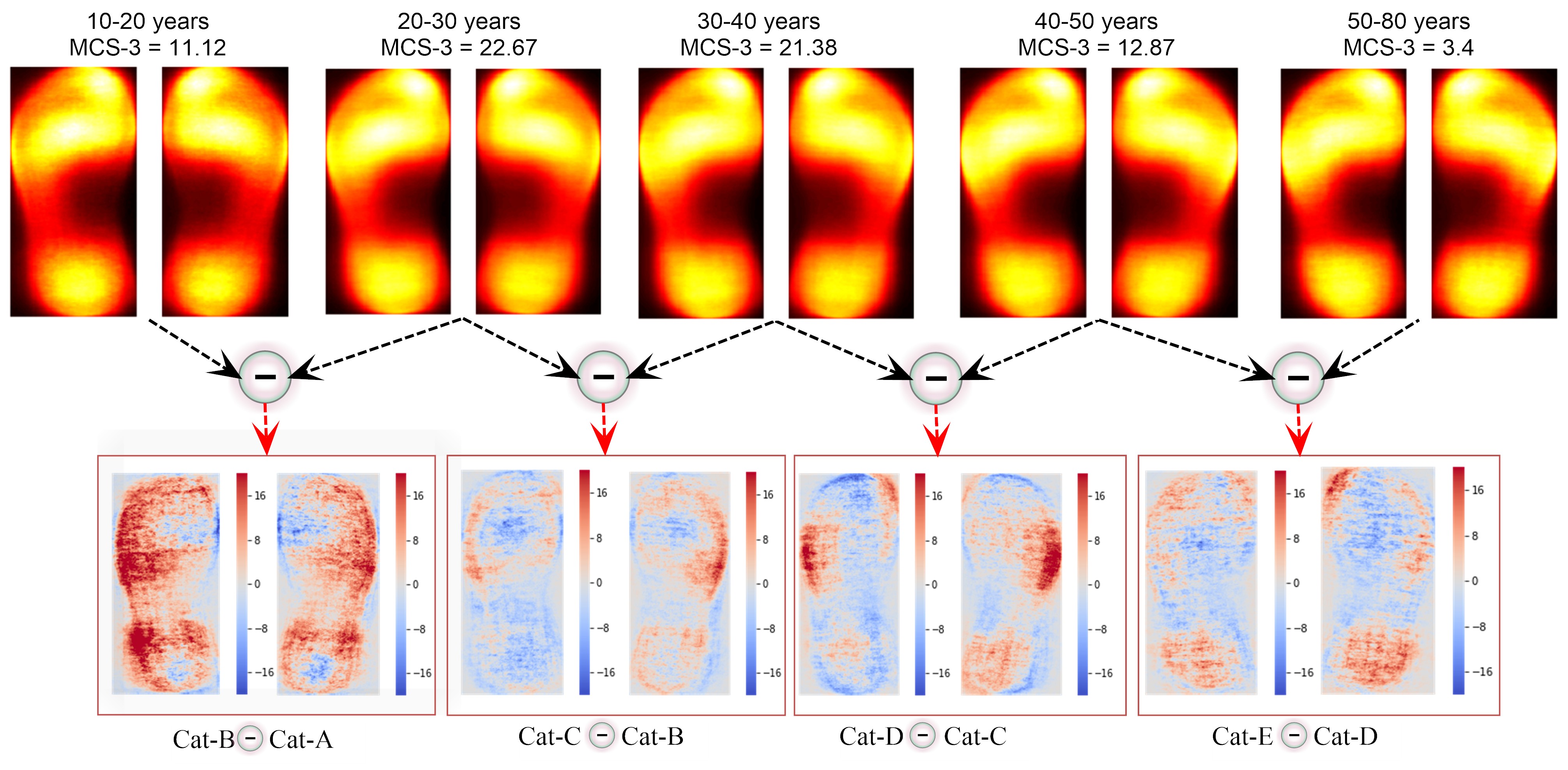}
	\caption{\normalfont{Category-wise pressure distributions on shoeprints with respect to age. The five age categories are shown with their corresponding \textit{MCS-3} values in the age prediction. The first row depicts the pressure distributions of the five-categories based on the given age ranges. All the visualized images are the average of superimposed shoeprints in each category. The second row renders the subtracted pressure differences between two neighboring categories.}}
	\label{fig:groupwise-subtraction}
\end{figure*}

\begin{figure*}[h!]
	\centering
	\subfigure[]{
		\raisebox{13mm}{\includegraphics[height=1.5in,width=.8in]{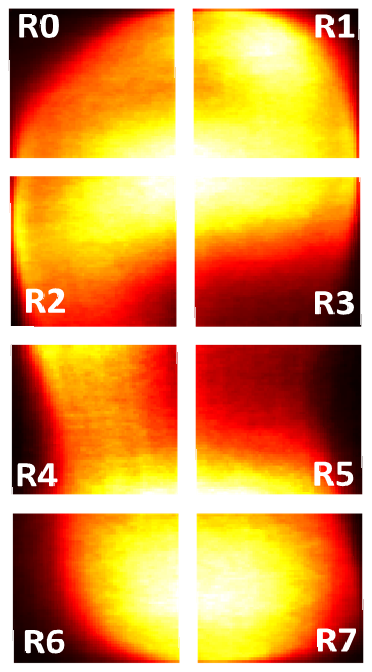}}
		\includegraphics[height=2.5in,width=2in]{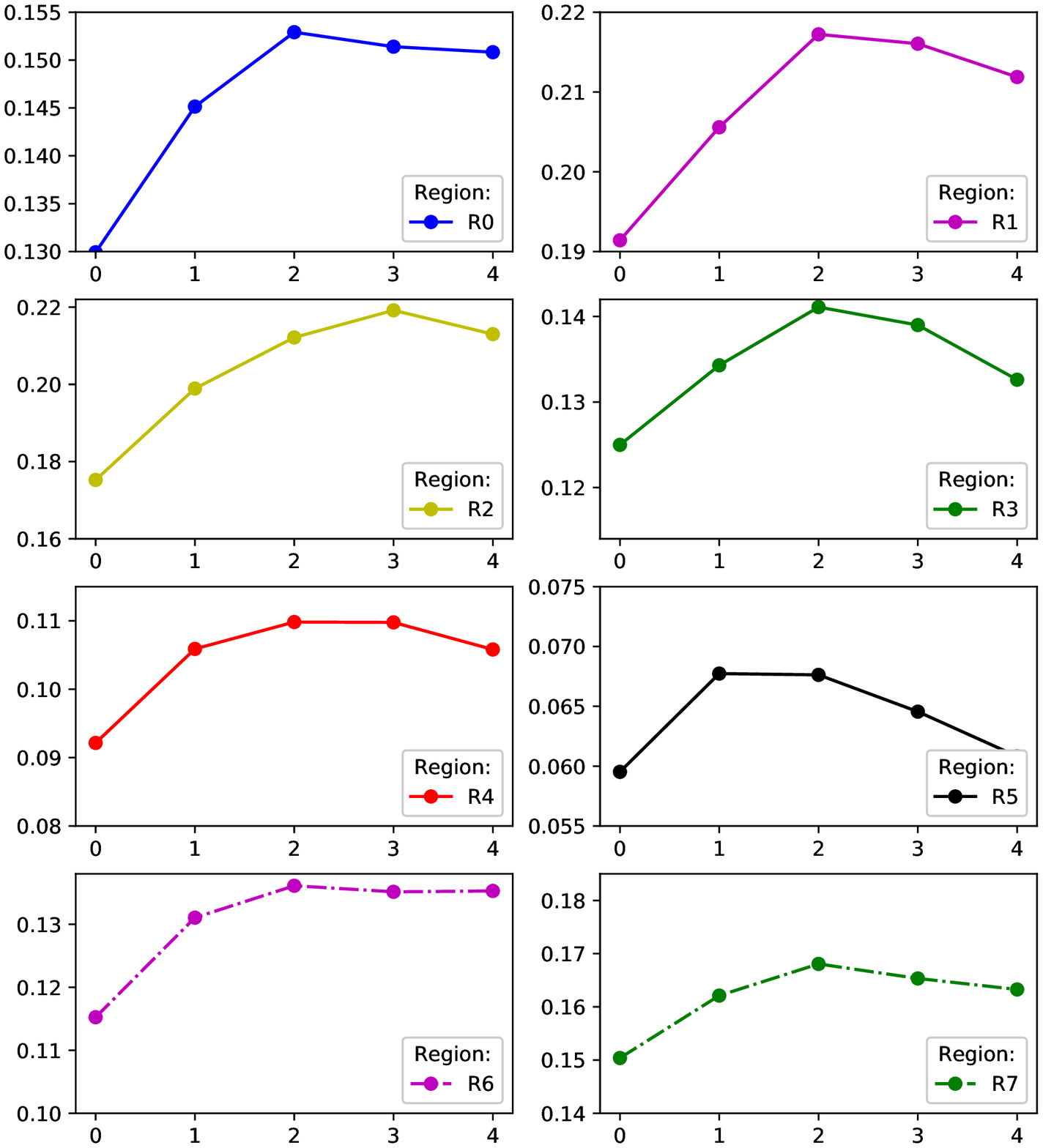}
		\label{fig:left shoeprint}
	}
	\subfigure[]{
		\includegraphics[height=2.5in,width=2in]{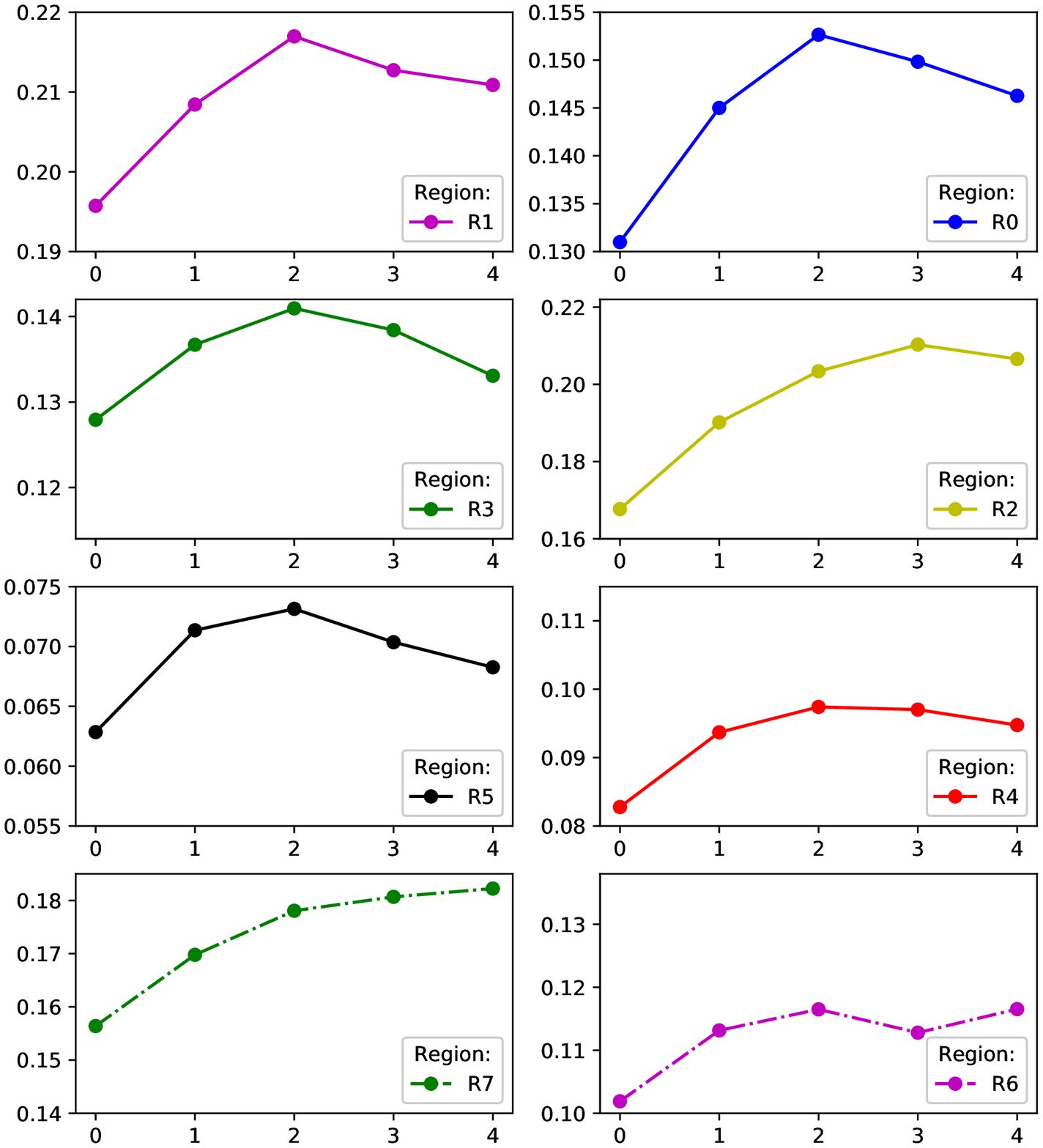}
		\raisebox{13mm}{\includegraphics[height=1.5in,width=.8in]{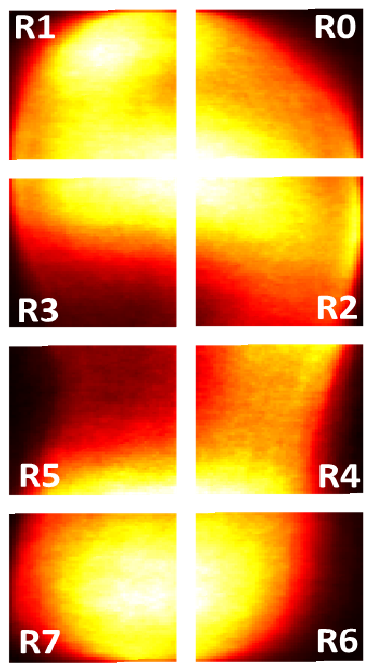}}
		\label{fig:right shoeprint}
	}
	\subfigure[]{
		\includegraphics[height=1.7in,width=2.9in]{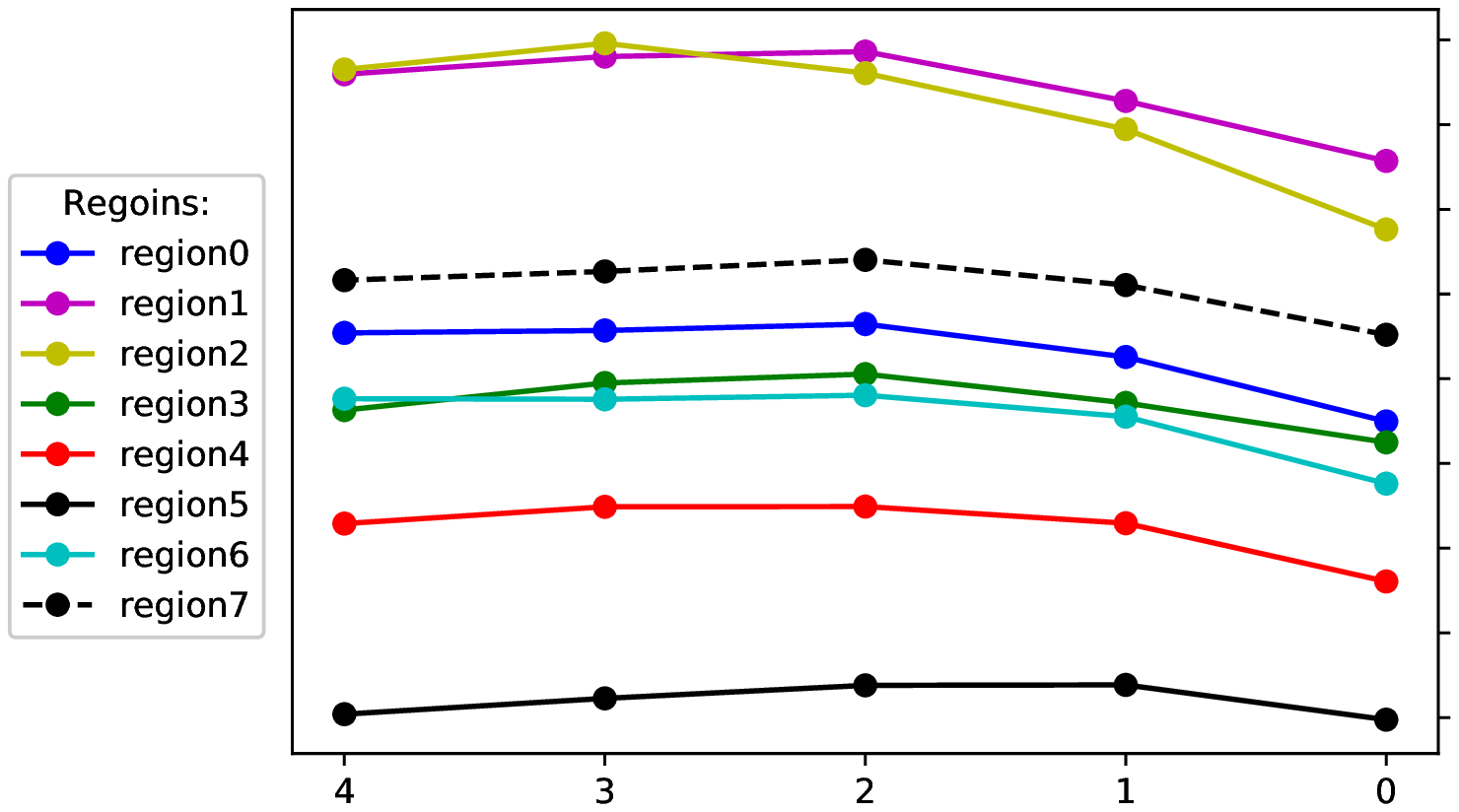}
		\label{ig:left curves}
	}
	\hskip -2.5ex
	\subfigure[]{
		\includegraphics[height=1.7in,width=2.9in]{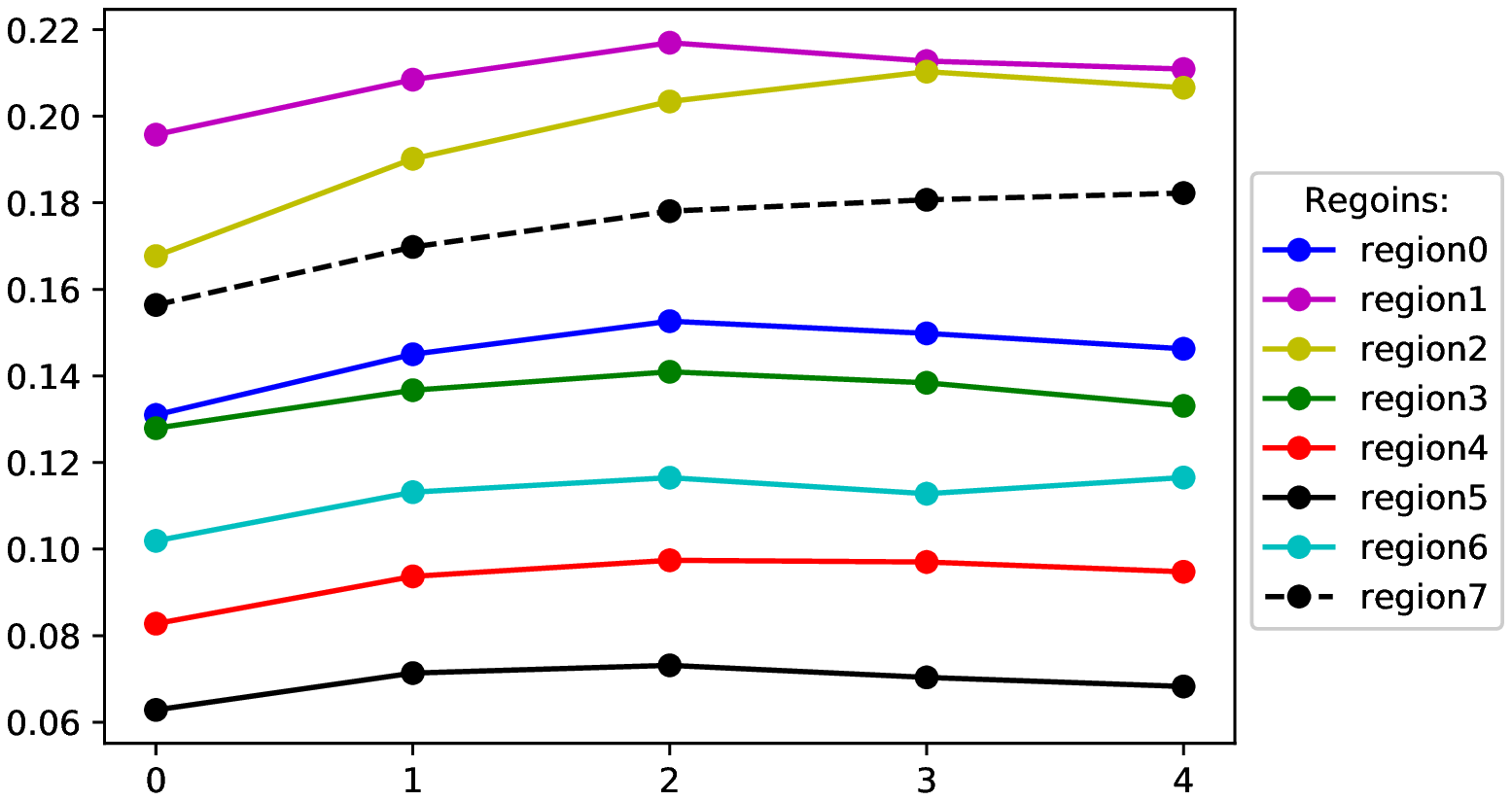}
		\label{fig:right curves}
	}
	\caption{\normalfont{The category-wise pressure distributions on shoeprints with respect to age for the 8-regions of L\&R shoeprints. In the first row, the left most and right most shoeprints are sampled from the average of superimposed shoeprints to visualize the divided eight regions. All the plots are demonstrated with pressure intensity values along the y-axis and aging along the x-axis. There are five age-based categorical division along the x-axis, with each group comprised of subjects according to their ages. For instance, Group 0 (Cat-A), Group 1 (Cat-B), Group 2 (Cat-C), Group 3 (Cat-C), Group 4 (Cat-D)  and Group 5 (Cat-E) along the x-axis represent ages ranging of 7 to 20, 21 to 30, 31 to 40, 41 to 50 and 51 to 80 years, respectively. All the shoeprints in their corresponding category are superimposed and then averaged to demonstrate the pressure trends. (a) and (b) show the category-wise aging effects in pressure distributions over eight divided regions in the given age ranges for left and right shoeprints, respectively. (c) and (d) show cumulative  trends of the pressure variations jointly on left and right shoeprints, respectively.}}
	\label{fig:eight-regions-curves-left-right}
\end{figure*}

To investigate the pressure distribution on shoeprints in standing and walking, an age group-based study can reveal the association of gait and standing-patterns with age. For this purpose, we used the division of Type-B in five categories (cat-A, cat-B, cat-C, cat-D, and cat-E) as mentioned above and shown in Table S7\cmmnt{~\ref{tab:tab3}}. All the shoeprints in each category were superimposed after segmentation and averaged along the given axis (see Methods). After the superimposition, L\&R shoeprints were obtained for each category. To depict the pressure trends, the corresponding subtraction process was performed for each category, i.e., the lower category subtracts from the higher category (e.g., cat-B--cat-A) along the given dimension, and this process continues for each category from that point on (see Methods). It can be deduced that the pressure distribution changes reflect pattern variations from one group to the next older group caused by aging, as shown in Figure ~\ref{fig:groupwise-subtraction}. The highlighted regions show that the front foot pressure is mainly on the first toe (hallux) as indicated in the first row of the Figure ~\ref{fig:groupwise-subtraction}, but it spreads toward other toes as shown in both the first and second rows. This pattern is also true for other areas, as the changes from (cat-B--cat-A) to (cat-E--cat-D) indicate that the high-pressure distributions shift to the outward regions of the heal with age progression. In particular, the first subtracted category (cat-B--cat-A) shows this category shift as the most distinctive, indicating obvious changes in pressure trends between the subjects of cat-A and cat-B. In the second superimposed subtraction (cat-C--cat-B), these pressures continue to shift toward the exterior regions of shoeprints but to a much lesser extent. Similarly, these pressures further move to the most exterior parts of shoeprints (cat-D--cat-C), with a slightly more pronounced shift toward the heel of the right shoeprints-moreso than toward the left shoeprints. Finally, the pressure shifts to the front toe area from cat-E to cat-D. These visualizations illustrate that the muscle forces reflected in shoeprints have a clear trend of variations from younger to elder ages.  
To further extrapolate the pressure distributions on shoeprints, the superimposed L\&R images of each category are divided into eight regions $(R_0, R_1, R_2...R_7)$ (Figure ~\ref{fig:left shoeprint}), and each region is depicted in curve trends as illustrated in Figure ~\ref{fig:right shoeprint}. These regions separately show the pressure trend variations with respect to age. The similar pressure distributions are portrayed when the upper and lower boundaries of shoes are aligned without the overall shown superimposition (Supplementary Figure S7\cmmnt{ ~\ref{fig:eight-regions-curves-left-right-all-same-scale}}). While in most areas, the patterns are symmetric between the left and right shoeprints, some clear symmetries are also observed. In the regions $R_0, R_1, R_3, R_4, R_7$, the intensities or pressure regions on both L\&R shoeprints spread increasingly from the early-age group (cat-A) to the middle-age group (cat-C) and then slightly decline trends in the elder-age group (cat-E) as shown in (Figure ~\ref{fig:left shoeprint}~\ref{fig:right shoeprint}). Noticeably, region $R\_2$ of both L\&R shoeprints has the most robust increasing trends of pressure over aging. A similar trend can also be noticed in Region $R_7$ of the right shoeprints but are not quite as noticeable in left shoeprints. The overall observation is that the pressure increases up to 40 years and steadily maintains on the outer part of the shoeprints (Figure ~\ref{fig:eight-regions-curves-left-right}(a,b)). The pressure intensities decline clearly after 40 years of age in regions $R_0, R_1, R_3$ and $R_5$ but to a less extent in $R_2, R_4, R_6$ and $R_7$. Overall, such a pressure decrease with aging in the inward regions is more pronounced than in the outward regions. The same pressure spread can also be seen in category-wise subtraction from the early-to-late ages (Figure ~\ref{fig:groupwise-subtraction}), while the subtractions of cat-A from the rest of the categories (cat-B, cat-C, cat-D, cat-E) are also visualized in Supplementary Figure S8\cmmnt{~\ref{fig:subtraction_of_early_category_from_the_rest_of_cat}}. With aging, the most visible changes are detected in the heel regions ($R_6,R_7$). Moreover, on left shoeprints, the pressure on region $R_2$ reaches or even exceeds $R_1$. All the pressure trends are cumulatively plotted for the corresponding left and right shoeprints along the age ranges (along the x-axis) as shown in Figure \ref{fig:eight-regions-curves-left-right}(c, d).  The cumulative plotting comparatively represents the evaluation of categorical age-ranges, which reveals variational effects in distinct periods of age groups.
\subsection{Pressure distributions between genders.} We analyzed the trends of pressure distribution changes for group-wise age groups based on gender. To substantiate the gender-based patterns, we proposed a gender classification network from shoeprints based on ShoeNet, which trains on Dataset-F (see Methods). The classifier predicts gender with a high accuracy (86.07\%), which infers obvious differences in patterns between males (M) and females (F) in pressure distribution. Males and females are found to vary in their gait patterns as they grow older growing. These variations can be seen for males and females in Supplementary Figure S9 and S10\cmmnt{\ref{fig:male groupwise pressure distribution subtraction} and \ref{fig:female groupwise pressure distribution subtraction}}, respectively, which shows males to have more pressure distribution toward the instep area than females. In females, the pressure varies significantly below 20 and above 40 years old. Similarly, male subjects show more pressure trends from early age to the age of 30 years; then, the pressure tends to flow smoothly toward shoeprint edges. In most cases, for both males and females, the right shoeprints are observed to endure more pressure forces with age. Similarly, the average (superimposed) image of every age group is divided into eight regions in the same manner as shown in Figure \ref{fig:eight-regions-curves-left-right}, which is visualized from early age (cat-A) to late age (cat-E) as shown in Figure ~\ref{fig:eight-regions-curves-gender-left-right}, based on male-female-left (MFL) and male-female-right (MFR) pressure related images. A large variable to aging is observed in the back-foot/heel region representing $R_6$ for females while the males do not have such a pattern. Region $R_2$ of MFL and MFR has some similar variations in the pressure range between 57.5 and 70.0 per unit area. Regions $R_2, R_4$ and $R_5$ of the corresponding MFL and MFR have similar patterns in pressure trends relative to aging. However, regions $R_4$ of both MFL and MFR, and $R_7$ of MFL have some opposite pressure trends with aging in both males and females. The overall analysis shows that males (red) have higher pressure trends than females (green) except in regions $R_0$ of both the left and right shoeprints, indicating a higher force used by females in this region, which is interesting. 
Collectively, we have a total of four sets of curves, each for male-left shoeprint (ML), female-left shoeprint(FL) (Figure ~\ref{fig:male_female_left shoeprint}) as well as for the male-right shoeprint (MR) and female-right shoeprint (FR) (Figure ~\ref{fig:male_female_right shoeprint}). From the depictions, we can observe that for region R5, lower pressures are common in ML and FL shoeprints. Furthermore, MR and FR shoeprints in regions R1 and R2 have higher pressures. Shoeprint regions $R_0, R_3, R_4, R_6$ and $R_7$ in ML have uniform variations in pressure trends while $R_0, R_3$ and $R_4$ of FL have similar distributions. In most cases, the pressure is higher for the male interior parts of L\&R shoeprints while females are only found to have higher pressure in the front toe ($R_0$). In conclusion, males and females have fluctuations in pressure forces in the form of shoeprints generated by gait and standing patterns with respect to age growth. Such fluctuations in pressure trends can be reasoned by body weight between males and females which are also reflected in terms of pressure distribution.
\section{Discussion and conclusion}
The main objective of this study is to estimate human age from imprinted shoeprints via a machine-learning approach. Shoeprints are influenced by the human gait and standing patterns, which are further attributed to aging. This is reflected in the shoeprint patterns for subjects under different age groups in different genders. We also demonstrated that ShoeNet can utilize such patterns to predict the human age and gender effectively. This study helps extend the frontiers of shoeprint applications from forensic investigations to diverse areas including biological profile estimations, examining health and sports activities, and clinical investigations. For example, in some cases, estimated physiological (biological) and actual (chronological) ages may not match well under different health conditions. For instance, some people look younger or older than their chronological age. Physiological aging could slow-down with a healthy environment, lifestyle, physical and cognitive functions\cite{121}. Predicted physiological age of an individual versus his or her actual age may provide some insight into the health conditions.
\begin{figure*}[h!]
	\centering
	\subfigure[]{
		\includegraphics[height=2.6in,width=3.1in]{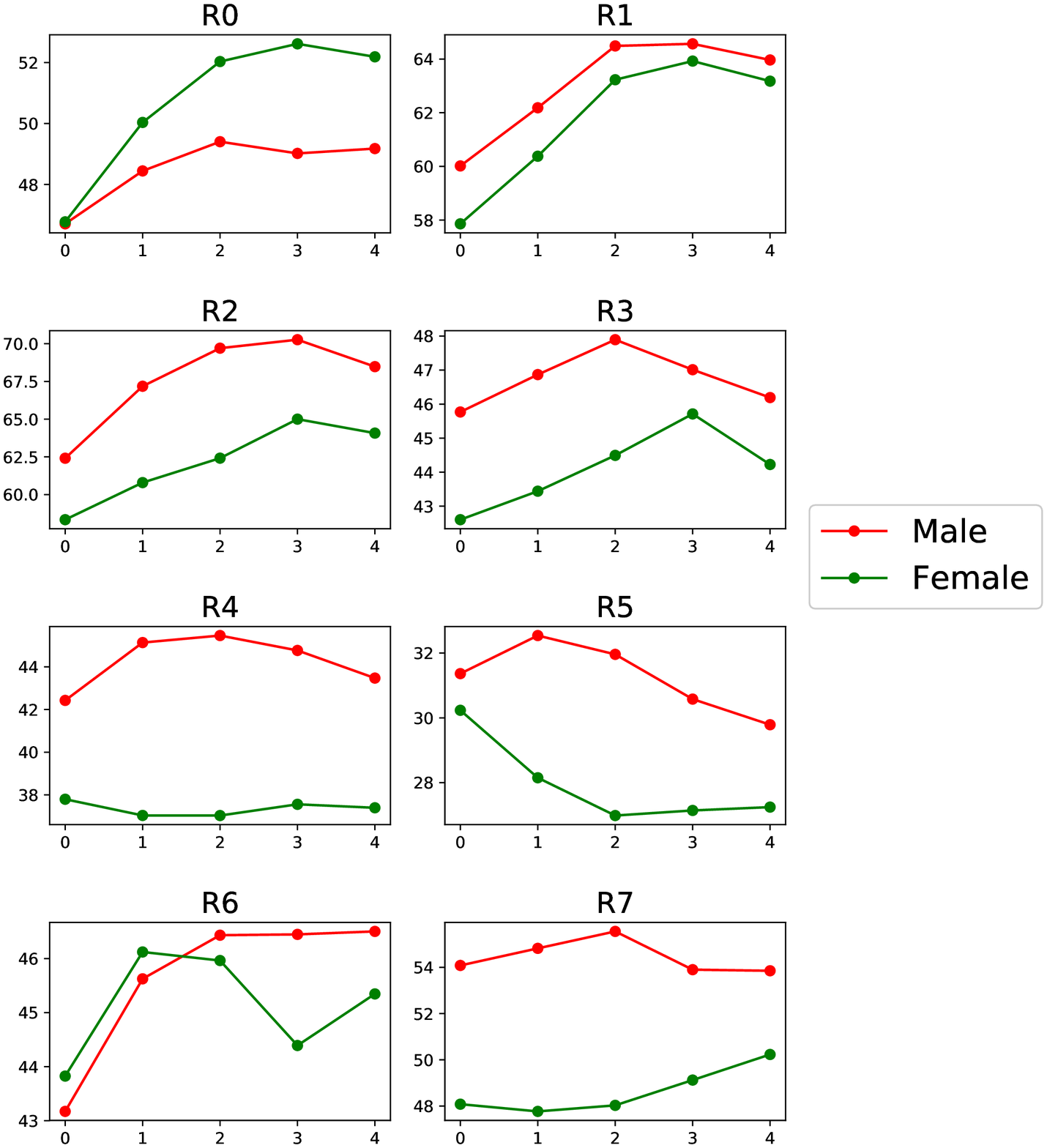}
		\label{fig:male joint female left}
	}
	\hskip -2ex
	\subfigure[]{
		\includegraphics[height=2.6in,width=2.8in]{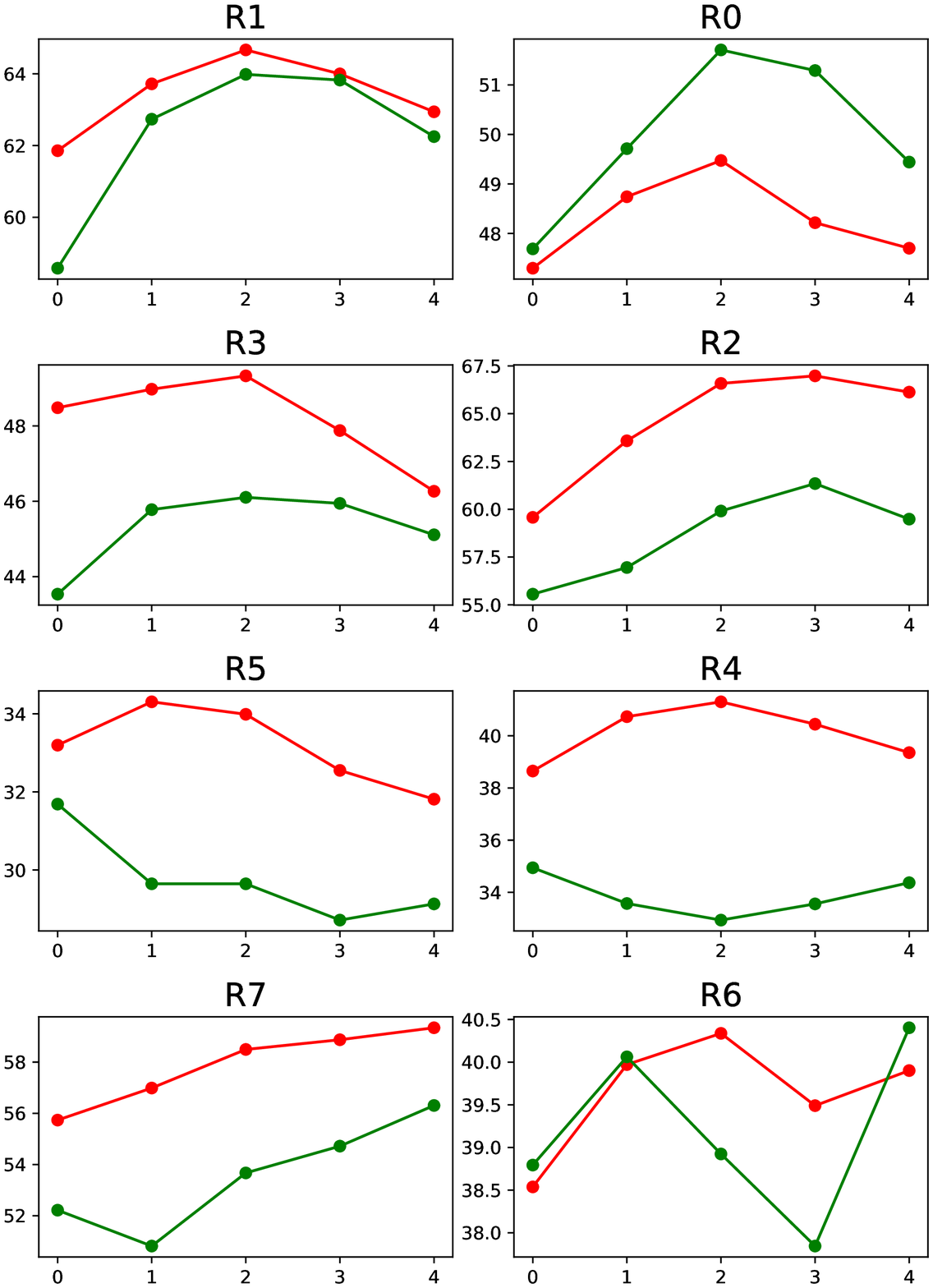}
		\label{fig:male joint female right}
	}
	\subfigure[]{
		\includegraphics[height=1.8in,width=3in]{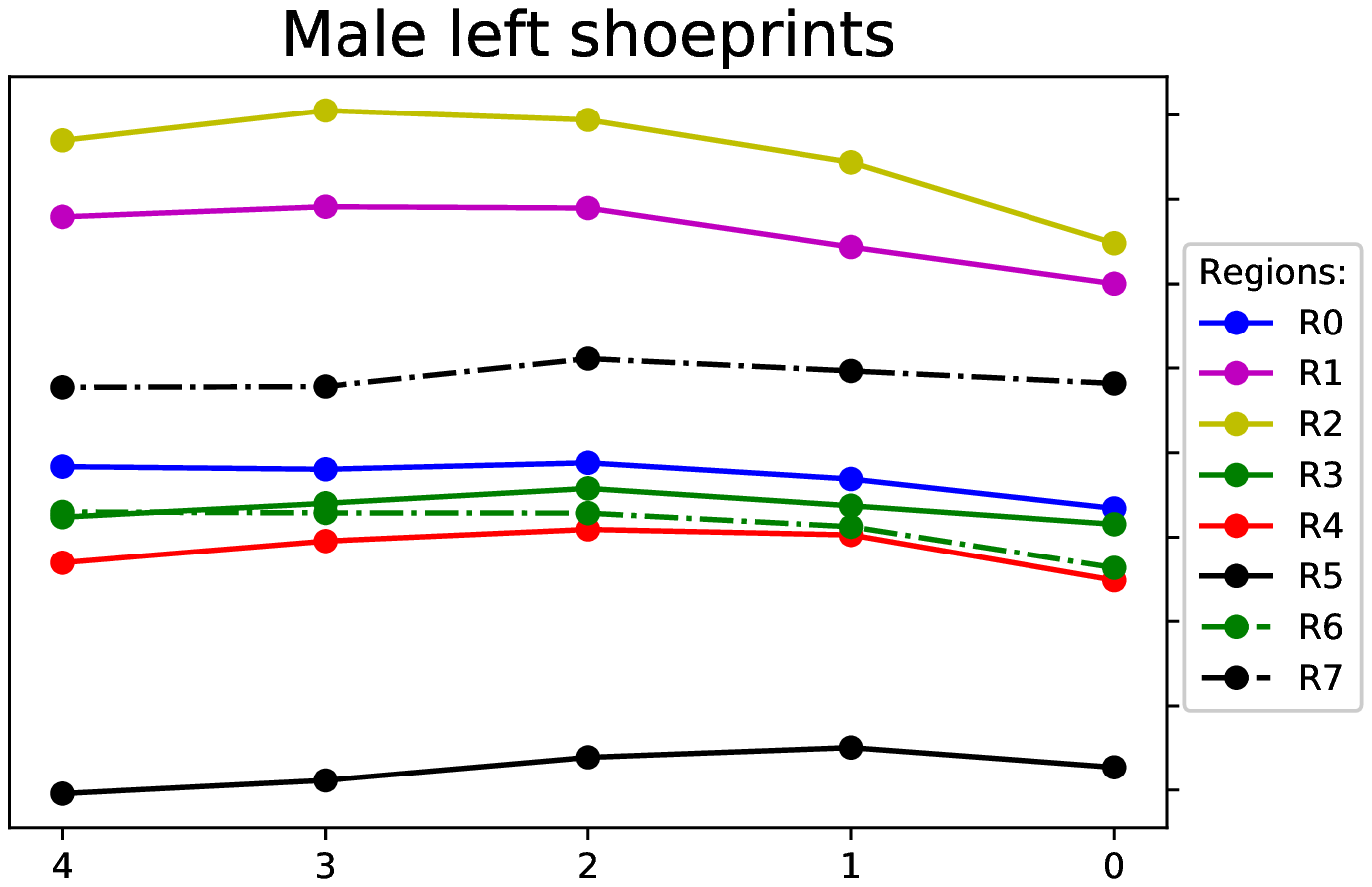}
		\hskip -1ex
		\includegraphics[height=1.8in,width=2.8in]{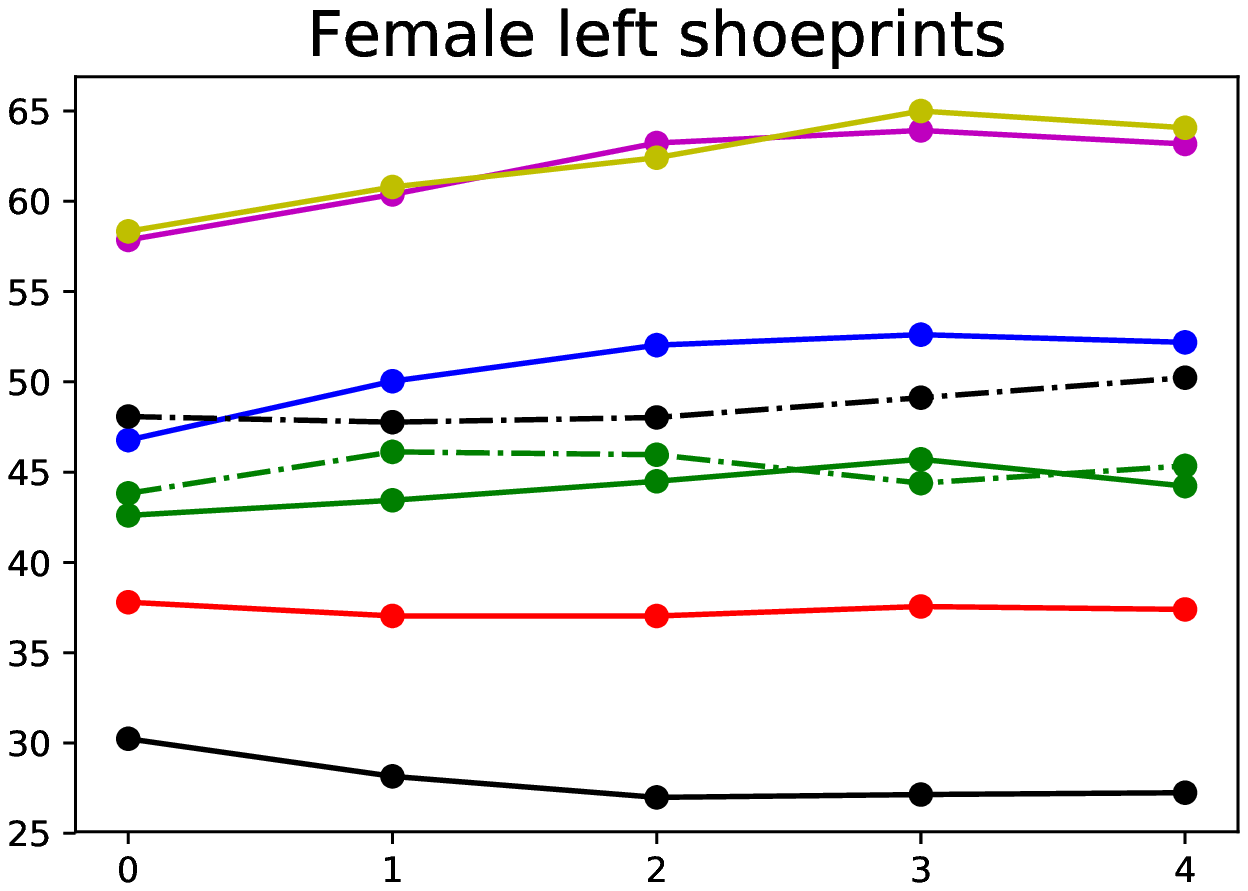}
		\label{fig:male_female_left shoeprint}
	}
	\subfigure[]{
		\includegraphics[height=1.8in,width=3in]{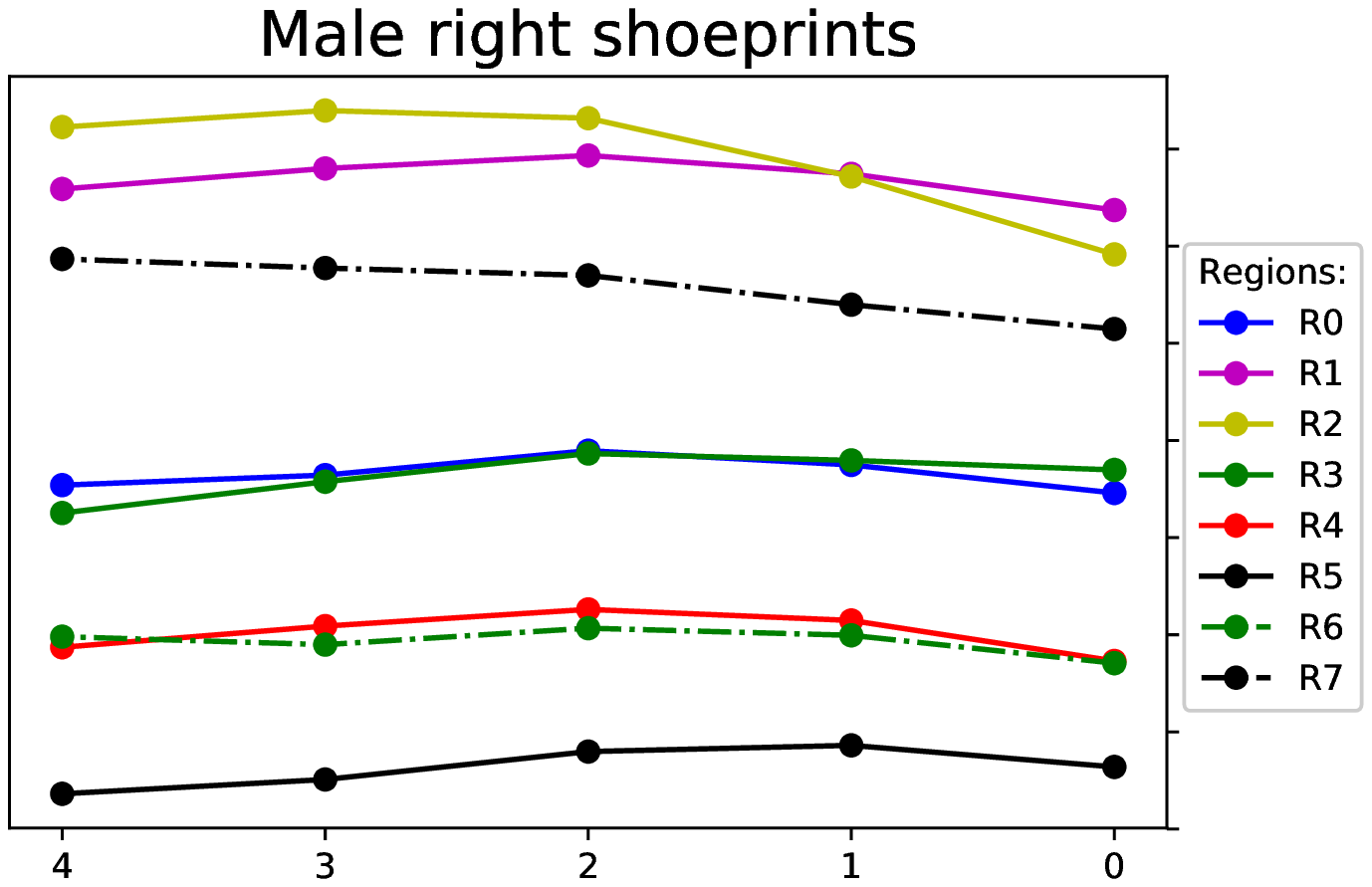}
		\hskip -1ex
		\includegraphics[height=1.8in,width=2.8in]{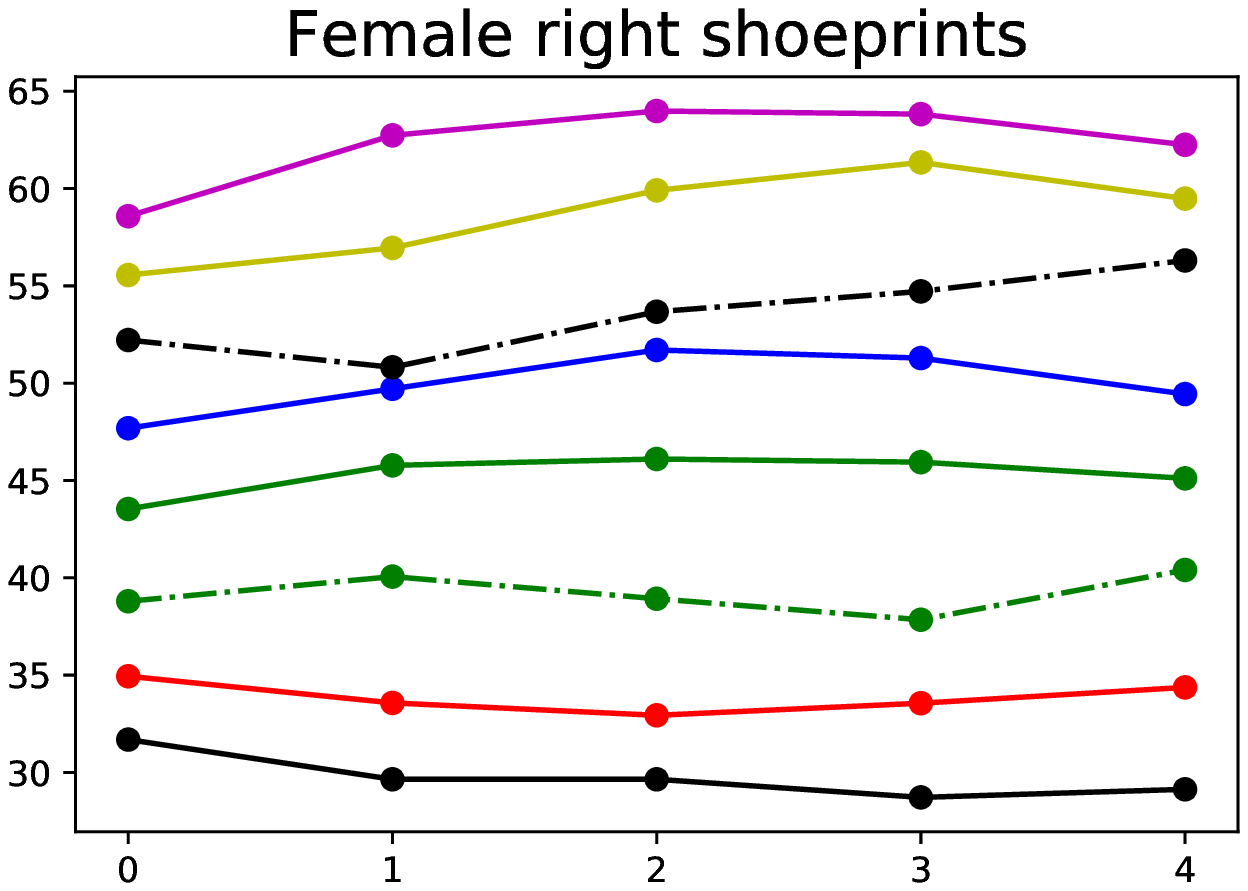}
		\label{fig:male_female_right shoeprint}
	}
	\caption{\normalfont{Category-wise aging effects in pressure distributions over eight divided regions between males and females, derived from the superimposed L\&R shoeprints of male-and-female (M\&F) in the given age ranges from cat-A to cat-E. Figures \ref{fig:eight-regions-curves-left-right} (a) and (b) show a similar scale for pressure intensities along the y-axis and for age groups along the x-axis, which cover the eight regions for M\&F for (a) left shoeprints and (b) right shoeprints. Each region has two curves. The red curve represents males, and green represents females. Collective trends are shown in (c) with male left shoeprints and female left shoeprints, and (d) with male right shoeprints and female right shoeprints. The relative regions with the same scale are shown in Supplementary Figure S11\cmmnt{ ~\ref{fig:eight-regions-curves-gender-left-right-same-scale}}
	}}
	\label{fig:eight-regions-curves-gender-left-right}
\end{figure*}
Age estimation is a challenging task even from facial, brain MRI\cite{92,93}, EEG\cite{95} and DNA\cite{94} due to many intrinsic and extrinsic factors including genetics, diet, hormones, and lifestyles. Technical challenges include insufficient labeled data and advanced computational methods. This study addresses both challenges by providing large-scale labeled data and a machine learning approach. It provides a benchmark for shoeprint-based age estimation by the availability of large-scale datasets and dataset-specific models. The prediction is based on the relationships among physiological aging, walking and gait patterns, and imprinted shoeprints. Our machine-learning model successfully addresses computational issues related to data noises, variational patterns, manufacturing designs, wear-time, and most importantly wear effects\cite{51,105}. In this study, the machine (deep) learning approaches are accomplished for age prediction within 5-years age ranges ($CS_5$). We had a good CS5 score of 40.23\%, and we also scored 86\% accuracy in classifying shoeprints in term of gender classification. It shows that the deep learning model learns knowledge from a large given dataset and extracts end-to-end features for aging and gender. The model is capable of capturing aging effects, which can be further improved by including more label information such as gender, height and weight. 

This study collected shoeprints from 50,000 subjects with annotations of age and gender, which derived a variety of datasets to be carried out in a wide range of experiments by training different machine-learning models to capture human gait and standing patterns related to aging  (Supplementary Table S6\cmmnt{~\ref{tab:tab2}}). Some of the trained networks are unable to capture gait patters. For instance, age prediction from the left or right shoeprints alone could not predict age accurately, as it may not capture much information about walking, standing and wear effects with diverse types of shoes, body weights, etc. In contrast, the proposed ShoeNet model trains on pair-wise shoeprints, and effectively captures features reflecting physical changes that appear with aging. This may indicate that age-related features are mostly reflected in the asymmetric differences between left and right shoeprints. ShoeNet adapts a deep convolutional neural network\cite{73,74} featuring extraction, and its architecture is designed to retain the pressure distribution varying with biological traits pertinent to aging, as well as the distribution differences between L\&R shoeprints. Another challenge of this study is a lack of consistent textual-outsole patterns in shoeprints with diverse types of shoes. ShoeNet utilizes pairwise shoeprints to fine-tune the network comprehensively and to capture areas of interest like scratches, cuts and abrasions with gait and standing patterns. ShoeNet retains morphological representations at a deep level capturing age related information, as well meaningful differences between left and right shoeprints related to aging. Hence, in a wide range of practices, ShoeNet outperformed most state-of-the-art methods based on the given datasets.

To explore aging effects on shoeprints in more detail, we grouped the subjects into a total of five age groups (Supplementary Table S7\cmmnt{~\ref{tab:tab3}}(Type-B)). Each group shows discrepant results for shoeprint-based age prediction. The subjects in the age group of 21-40 years old have the most consistent pressure distribution on shoeprints reflecting the robust association of age to gait and standing patterns while the rest of the subjects vary more due to body and health conditions. Similarly, body morphological effects on shoeprints can be illustrated in the form of a pressure-spreading anatomy (Figure ~\ref{fig:groupwise-subtraction}). The most obvious pressure trends were found spreading from the interior to exterior regions over aging. In particular, the pressure trends increase up to 40 years old and then a steady decline is observed which was reflected in regions $R_0, R_2, R_3$ and $R_5$ (Figure \ref{fig:eight-regions-curves-left-right}(a, b)). Previous studies have shown that abrasion points occur with the specific pressure of foot-muscles on certain areas of shoeprints while in contact with the ground surface\cite{2,31}. Our study provides evidence for such intensive forces on shoeprints with the depiction of pressure trends (Figure ~\ref{fig:eight-regions-curves-left-right},~\ref{fig:eight-regions-curves-gender-left-right}). We also expanded the frontier of gender-based wear effects on shoeprints in the capacity of pressure distributions (Figure ~\ref{fig:eight-regions-curves-gender-left-right}), which may assist in a wide range of fields, especially forensic podiatry. Similarly, in earlier studies, shoeprint-based gender predictions were carried out from morphological features (like shoeprint length and width) and via regression by using hand-crafted features\cite{117,118,119}, while ShoeNet adopts convolution filters for feature extraction from the pictures of shoeprints directly for the end-to-end gender prediction (86.07\%) in the large-scale balanced dataset (Supplementary Table S4\cmmnt{~\ref{tab:gender-testing-result}}). We can observe that the males and females have distinct pressure distributions in terms of intensity values, trend patterns, and age growing fluctuations in the corresponding regions. Evidently, males have higher pressure trends except for the outer front toes of both L\&R shoeprints (region $R_0$) (Figure ~\ref{fig:eight-regions-curves-gender-left-right}(a and b)). The high gender classification accuracy (86.07\%) also confirms study findings about gait pattern variations between males and females\cite{53}. Taken together, ShoeNet exploits varying effects in distinct age groups and achieves good result for age estimation and gender prediction. This study will provide useful information and methods for forensic investigations, gait-pattern disorders associated with aging, and biological profile estimations, as well as sports and clinical investigations.
\section*{Acknowledgments}
This research is supported by grants to YZ, including the National Natural Science Foundation of China (Grants Nos. 61772227, 61972174, 61972175), the Science and Technology Development Foundation of Jilin Province (No. 20180201045GX, 20180414012GH), and the Jilin Development and Reform Commission Fund (No. 2020C020-2), as well as the Paul K. and Diane Shumaker Endowment Fund to DX. We like to thank Ms. Carla Roberts for thoroughly proofreading this paper. 
\section*{Data availability} 
All the initial datasets were collected via the EverOS V2.0 acquisition system by Everspray Science and Technology Company Ltd., Dalian, China, and the rest of the dataset versions were generated by manual and automatic processing. All the datasets are available to the research community following the terms and conditions set forth at http://www.everspry.com/en/service/data\_open.htm. 
\section*{Code availability} 
The code and related documents are available on https://github.com/mhassandev/ShoeNet.

\ifCLASSOPTIONcaptionsoff
  \newpage
\fi



%

\bibliographystyle{IEEEtran}

\bibliography{Main}
	\newcommand{\beginsupplement}{%
	\setcounter{table}{0}
	\renewcommand{\thetable}{S\arabic{table}}%
	\setcounter{figure}{0}
	\renewcommand{\thefigure}{S\arabic{figure}}%
}
\beginsupplement
\onecolumn
\section{ShoeNet-based comparative study.}
We computed the effectiveness of ShoeNet by comparison with the state-of-the-art and custom-network modalities. In both the studies, ShoeNet outperforms standard modalities (main section Figure 3\cmmnt{~\ref{fig:Comparative-study-graph}}) as well as custom modalities (Table ~\ref{tab:tab2}), which contains a list of dataset versions generated to estimate shoeprint-based profile traits. We customized deep learning modalities using the list of datasets to investigate the effects of age and gender on shoeprints. For this purpose, we carried out three core deep modalities (LR-CNN, FM-CNN, MM-CNN), which are further divided into sub-modalities.
\subsection{Left-to-right CNN (LR-CNN).} In this experiment, the model was trained on both L\&R shoeprints to carry out age estimation and their corresponding association to gait patterns. Our model has the same structure as ShoeNet (depicted in Figure 2\cmmnt{ ~\ref{fig:suggested-modde}}), we input distinct dimensions (224$\times$112) of left and right shoeprints separately instead of using one combined image containing both left and right shoeprints (which is the practice in ShoeNet). In the study of LR-CNN, three different datasets, (a) left-shoeprint (Dataset-C), (b) right-shoeprint (Dataset-D) and (c) L\&R shoeprints (C+D) were used to train and test LR-CNN. The corresponding datasets for (a), (b) and (c) are depicted in Supplementary Figure ~\ref{fig:Unprocessed_dataset} and a detailed description can be found in Supplementary Table ~\ref{tab:datasets}. The network deployed he skip-layers concept learning wear patterns from shoeprints. The three possible datasets, described as left-shoeprint (only left shoeprints for training and testing), right-shoeprint (only right shoeprints for training and testing), and L\&R (including both L\&R shoeprints), were used to extrapolate the association of aging with gait and standing-patterns. The network merges the skip layers to the next deep-level layers to learn wear patterns from shoeprints. The mean-absolute-errors (MAE) and the \textit{MCS-J} scores for (a) left-to-right networks are shown in Supplementary Table ~\ref{tab:tab2}. The learning curves of the three subnetworks are depicted in Figure ~\ref{fig: LR-CNN}(a). Similarly, their corresponding validation error curves can be seen with the lowest error found in the left-shoeprint during training and validation (Supplementary Figure ~\ref{fig: LR-CNN}(b)). It can also be observed that the networks based on left-shoeprints and on right-shoeprints learn speedily as compared to the network having both mingled shoeprints i.e. L\&R. Similarly, the age prediction is higher (with 12.12 \textit{MCS-2} score) for the right-shoeprints among the three observed modalities and datasets as shown in Table ~\ref{tab:tab2}.
\subsection{Effectiveness of Fusion modeling}. The fusion network modalities at different levels have their pros and cons, which vary corresponding to the nature of the deep learning problems. In order to study both the left and right-shoeprints simultaneously, CNN-based fusion models integrate at different levels to investigate the learning process, feature extractions, computational power, and score prediction for age estimation. Similarly, the fusion model (FM-CNN) receives shoeprints in the same dimensions as received by LR-CNN $(width = 112, height = 224)$ despite dissimilarities in the network structure. The fusion is carried out between two identical networks at the input-level, mid-level and decision level to improve age prediction from shoeprints. In fusion models, the merging performs at Block-A, Block-C, and Block-E considering the effectiveness in the corresponding network. The structure of each block contains convolution layers, batch normalizations, and ReLu activations followed by a max-pool-2d layer. In the process of merging, the features concatenate element-wise along the third dimension. In the early fusion, we merged the information into a stack of features varying from left to right shoeprints in the early layers of the network as described in Supplementary Figure ~\ref{fig:Fusion Model}(a). Different feature vectors keep discriminative information extracted for the same pattern at distinct inputs while discarding redundant information. Here, the multiple feature vectors considers for a pair of shoeprints i.e. left-shoeprint and right-shoeprint. Both L\&R shoeprints are accepted into two twin networks for the extraction of low-level features to optimize the deep learning process as it relates to age. Both twin networks merge at level-A into a single network model for further feature extraction. In contrast to the early fusion, the in-fusion accomplishes at the mid-level of both networks after extracting features from deep layers (Supplementary Figure ~\ref{fig:Fusion Model}(b)). The features are combined into a single stack of feature-representation after incorporating the blocks (Blocks-A, B, C) for finding effectiveness at mid-level Figure-~\ref{fig:Fusion Model}(b). It also directs the networks to learn from the features extracted in isolation as well as in combined form. Similarly, late fusion merges the results of multiple networks into a single representation at the decision level prior to the fully connected layers (Supplementary Figure ~\ref{fig:Fusion Model}(c)). The corresponding training and validation curves for early-fusion, in-fusion and late-fusion networks are shown in Supplementary Figure ~\ref{fig:fusion-left-to-right_training_validation}. All the fusion models are more costly in number of computational operations, while among them, in-fusion produces better results in training and validation (Supplementary Figure ~\ref{fig:fusion-left-to-right_training_validation}) as well as in testing (\textit{MCS-2}=11.19) (Table ~\ref{tab:tab2}).
\subsection{Multi-Model Parameters Sharing.} CNN-based multi-model (MM-CNN) imitates the concept of a Siamese neural network\cite{96}, which enables sharing multi-level information hierarchically to delve into the age estimation process. To establish information sharing, the models receive both left and right shoeprints simultaneously to extract features at different hierarchy of filters, thereby strengthening age prediction. The identical subnetworks share common weights to capture salient features at the underlined positions as illustrated in Supplementary Figure  ~\ref{fig:multi-model}. Parameter sharing (PS) accomplishes the transferring and sharing of learned weights between the twin networks at level-A (Early-MM), level-B (Mid-MM) and level-C (late-MM). MM-CNN takes L\&R shoeprints as the input to both identical CNN based networks and down-sample the features at a particular level; and then, MM-CNN learns multi-model embedding to excite the generalization toward age estimation. The enfolding and concatenation of all features perform at the lower level prior to the fully connected (FC) layers. In the PS between the twin networks, the FC layers mitigate the feature size of perceptive filters to merge at a latent dimension. The merging performed channel-wise has a structure similar to the computed mean and standard deviation. The latent representation during squeezing is a summation of concatenated filter-wise and random-normal (Gaussian) based value. The resulting value is then summed up filter-wise to excite and recalibrate the input features. This joint representation allows one modality to recalibrate the features in another modality. For instance, the features obtained from one modality having a significant impact would excite the same features to optimize in the other modality. Thus, both network modalities utilize the parameters sharing and tuning at different hierarchal levels. In the empirical results of MM-PS, early-MM has a fast convergence rate compared to mid-MM and late-MM regarding training (Supplementary Figure ~\ref{fig:multi-model-sharing-training-and-validation}(a)) and validation (Supplementary Figure ~\ref{fig:multi-model-sharing-training-and-validation}(b)) as well as a higher $MCS-J (11.09, 14.59 for J=2,3)$ score as illustrated in (Table ~\ref{tab:tab2}). The learning and error rates for all the modalities are the same for as those $20,000$ steps. Afterward, the error rate for early-MM drops significantly, achieving a better result.
\clearpage
\newpage
\onecolumn
\begin{table*}[h!]
	\caption{Seven dataset versions.}
	\label{tab:datasets}
	\footnotesize		
	\centering
	\renewcommand{\arraystretch}{1.3}
	\setlength{\extrarowheight}{1pt}
	\begin{tabular}{l|p{2cm}|p{2.5cm}|c|c|c|c|c|p{3cm}}
		\hline Dataset Name&No. of images& Image size (HxW) & \centering{LSO*} & \centering{RSO*}&\centering{BLR*}&Ruler&Gender&Description\\ \hline
		Dataset-A &100,000 &224x112 &\centering{-} &\centering{-}&\centering{\checkmark}&\checkmark&-&Original dataset\\
		Dataset-B&100,000	& 224x112 &\centering{-}&\centering{-}&\centering{\checkmark}	&-& -&Ruler-less and flipped dataset	\\
		Dataset-C&42,890 &224x112&\centering{\checkmark} &\centering{-} &\centering{-}&-& -&Left shoeprints dataset	\\
		Dataset-D&42,890& 224x112 &\centering{-} &\centering{\checkmark} &\centering{-}&-& -&Right shoeprints dataset\\
		Dataset-E&42,890	& 224x224 &\centering{-} &\centering{-} &\centering{\checkmark} &-&-&Pairwise dataset	\\
		Dataset-F&60,482	& 224x112 &\centering{-}&\centering{-}&\centering{\checkmark}	&-& \checkmark& Gender dataset \\
		Dataset-G&151,000	& 224x112 &\centering{-}&\centering{-}&\centering{\checkmark}	&-& -&Balance augmented dataset	\\
		\hline
	\end{tabular}
	\\
	\justify *Left shoeprints only-LOS, *Right only-RSO, Both left-and-Right-BLR. The list of dataset versions was generated to practice a wide range of experiments related to age estimation. Each dataset is described by number of samples, image-dimensions, whether including left or right or both shoeprints, scale, and gender information. The checkmarks $(\checkmark)$ exhibit the existence of the features. Dataset-C and Dataset-D were used for the comparative study of custom modalities, while standard modalities and ShoeNet were trained on Dataset-G. Dataset-G was augmented to balance the number of samples per age period. Similarly, Dataset-F was used for shoeprint-based gender classification and analysis.
\end{table*}
\begin{table*}[!h]
	\caption{Data distribution of males and females in the original and augmented datasets.} 
	\label{tab:gender-dataset-distribution}
	\footnotesize		
	\centering
	\renewcommand{\arraystretch}{1.3}
	\begin{tabular}{l|>{\raggedleft}p{1cm}|p{1.5cm}|c|c|c|c|c}
		\hline\multirow{2}{*}{S.No}&\multicolumn{2}{c|}{Dataset} &\multicolumn{2}{c|}{Augmented}&\multicolumn{2}{c|}{Total} &Grand Total\\ \cline{2-8}
		&Male &Female & Male&Female&Male&Female&Male+Female\\ \hline
		Original&24371&5120&-&-&-&-&29491\\
		Training&22871&3620&-&21100&22871&24720&47591\\
		Validation&-&-&-&-&-&-&10\% of 47591\\
		Testing&1500&1500&-&-&-&-&3000 of 29491\\ \hline
	\end{tabular}
	\\
	\justify Male subjects have much more samples than female subjects. To address the imbalance, the female samples were augmented from 5120 to 22,871. So, a total of 47,591 samples were used for training, in which 10\% were used for validation, and 3000 of the original samples were used for testing.
\end{table*}
\begin{table*}[h!]
	\caption{Comparative evaluation-scores of ShoeNet trained with Custom-Loss-Function (CLF) and with MSE loss functions.}
	\label{tab:CLS-vs-MSE}
	\setlength\tabcolsep{3pt} 
	\footnotesize		
	\centering
	\renewcommand{\arraystretch}{1.3}
	\begin{tabular}[h!]{l|c|c|c|c|c|c|c}
		\hline Network&\%MAE	 & $CS_0$ & $CS_1$ & $CS_2$&$CS_3$&MCS-2&MCS-3\\ \hline
		CLF&11.20& 4.3&13.14&20.69 &27.09&12.71&16.305\\ 
		MSE&11.07& 4.04 &12.84&19.49 &25.73 &12.12 &15.52\\
		\hline
	\end{tabular}
	\\
	\justify CLF was customized for ShoeNet to predict age, in which the loss values out of the range are treated with more penalization than the in-range values. The percent accuracy of CLF-based ShoeNet is less accurate than the MSE-based trained ShoeNet. Similarly, the cumulative scores including \textit{MCS-J} $(for J=2,3)$ are also significant for ShoeNet model trained with the CLF loss function. CLF $(for j=2,3)$ and MSE based loss values are visualized in Figure ~\ref{fig:CLFvsMSE}
\end{table*}
\begin{table}[!ht]
	\caption{Gender-based classification report with significant (86.07\%) accuracy.}
	\label{tab:gender-testing-result}
	\footnotesize		
	\centering
	\renewcommand{\arraystretch}{1.3}
	\begin{tabular}{c|c|c|c}\hline
		&Precision&Recall&F1-score \\ \hline
		Male&0.8360&0.8973&0.8656\\
		Female&0.8892&0.8240&0.8554\\ \hline
		Testing accuracy&\multicolumn{3}{c}{86.07\%}\\  \hline
	\end{tabular}
\end{table}
\begin{table*}[h!]
	\caption{Comparative study of the ShoeNet with the standard deep learning models}
	\label{tab:Comparative-study}
	\setlength\tabcolsep{11pt} 
	\footnotesize		
	\centering
	\renewcommand{\arraystretch}{1.3}
	\begin{tabular}{l|p{2.5cm}|c|c|c|c|c|c}
		\hline Network&(\%)Accuracy MAE	 & $CS_0$ & $CS_1$ & $CS_2$&$CS_3$&MCS-2&MCS-3\\ \hline
		RandomMd &6.88&0.95&	3.35 &6.50&	9.10&3.56&4.97\\
		Highway Net &6.88&1.99&6.44 &10.64&	15.19&6.36& 8.565\\
		DenseNet &10.44	& 2.29 &7.65&13.04&18.94	&7.66	&10.48	\\
		AlexNet&9.64&2.64 &9.54&15.19&20.83	&9.12&12.05	\\
		VGG16 &9.98	& 3.49 &10.49 &17.34 &23.28	&10.44&13.65\\
		GoogleNet&\textbf{11.20}&\textbf{4.93}&11.84 &18.44 &24.58&11.74&14.95\\
		Inception V4&10.66& 4.04 &11.19 &17.04 &24.08&10.76&14.09\\
		VGG19&10.88	& 3.84 &12.64 &18.84 &25.93 &11.78 &15.31\\
		\textbf{ShoeNet}&10.90&4.64&\textbf{13.79}&\textbf{20.73} &\textbf{28.03}&\textbf{13.06}&\textbf{16.80}
		\\
		\hline
	\end{tabular}
	\\
	\justify The bold font indicates the best performance in the category. RandomMd randomly took samples in the given age ranges (10-20). The evaluation metrics Percent-MAE, $CS_j (for j=0,1,2,3..n)$ and \textit{MCS-J} $(for J=2,3)$ are used to examine the significance of under-studied network modalities. For percent-MAE and $CS_0$, GoogleNet achieves the best result (11.20) while ShoeNet has the highest scores for cumulative metrics of \textit{MCS-2} (13.06) and MCS-3 (16.80).
\end{table*}
\begin{table*}[h!]
	\caption{Four main network modalities and their corresponding values for MAE, MCS-2 and MCS-3.}
	\label{tab:tab2}
	\setlength\tabcolsep{20pt} 
	\centering
	\renewcommand{\arraystretch}{1.3}
	\begin{tabular}{p{0.15cm}|p{2cm}|p{4cm}|p{0.3cm} p{0.9cm} p{0.9cm}}
		\hline
		No.&Networks &Network-Types & MAE & MCS-2 & MCS-3\\
		\hline
		1.&&\textbf{a}-Left-Shoeprints &9.38&	11.52&15.16\\
		&LR-CNN&\textbf{b}-Right-Shoeprints&9.70&12.12&15.70\\
		& &\textbf{c}-Left-to-Right Shoeprints   &9.51&10.49&13.78\\ \hline
		2.&&\textbf{a}-Early-Fusion	&9.48&10.19&13.60\\
		& FM-CNN&\textbf{b}-In-Fusion	&9.78&11.19&14.74\\
		& &\textbf{c}-Late-Fusion    &9.45&9.91&13.15\\ \hline
		3.& &\textbf{a}-Early-Sharing&8.99	&11.09&14.59\\
		& MM-CNN&\textbf{b}-Middle-Sharing	&9.10&10.74&14.31\\ 
		& &\textbf{c}-Late-Sharing	&9.72&9.96&13.35\\  \hline
		4. &ShoeNet&ShoeNet&\textbf{9.21}&\textbf{13.06}&\textbf{16.80}
		\\
		\hline
	\end{tabular}
	\\
	\justify The four main network modalities are customized using the generated datasets for training. Each modality uses left and right shoeprints for training while having a distinct way of processing. The four modalities are further divided into sub-networks having distinct internal structures. In the first category, the right shoeprints-based network has a higher score. Similarly, in the fusion modalities of FM-CNN, In-fusion has a significantly better result than early and late fusions. Furthermore, the early sharing model has a significantly better result than middle or late sharing. Overall, ShoeNet has the highest cumulative score and outperforms the other modalities, although the mean absolute value is not significant for ShoeNet as a result of customization in the loss function leveraging evaluation metrics (MCS-J).
\end{table*}
\begin{table*}[h!]
	\caption{Statistical results for age estimation in different age groups.} 
	\label{tab:tab3}
	\setlength\tabcolsep{16pt} 
	\footnotesize		
	\smallskip 
	\centering
	\renewcommand{\arraystretch}{1.3}
	\begin{tabular}{c|c|c|c|c}
		\hline {No.}&{Age ranges in years}&{MAE mean-absolute-error} &{MCS-2}&{MCS-3} \\ \hline 
		\multicolumn{5}{c}{\textbf{Type-A}} \\ \hline
		1.&10-80& 9.21&13.06& 16.80\\
		2.&20-50&7.51&15.16 &19.55 \\
		3.&25-45&6.44&17.59&22.00\\
		\hline\multicolumn{5}{c}{\textbf{Type-B}} \\ \hline
		4.&10-20&10.80&9.24	&11.12\\
		5.&21-30& 6.77&17.86&22.67\\
		6.&31-40&6.23  &16.52&21.38\\
		7.&41-50&10.05&9.64&12.87\\
		8.&51-80&18.82& 2.4 & 3.4\\ 
		\hline
	\end{tabular}
	\\
	\justify  Type-A has three subtypes while type-B has 5 subtypes based on the age ranges. Type-A has the best MAE (6.44) and \textit{MCS-J} $(17.59 for J=2)$ scores in the age group ranging from 25-45 years old. Similarly, Type-B has the best score of MAE (6.23) for the age group of 31-40 years old while category 21-30 has the best \textit{MCS-J} score $(17.86 for J=2)$.
\end{table*}

\newpage
\clearpage
\section*{Supplementary Figures}
\begin{figure*}[h!]
	\centering
	\includegraphics[height=3in,width=3.5in]{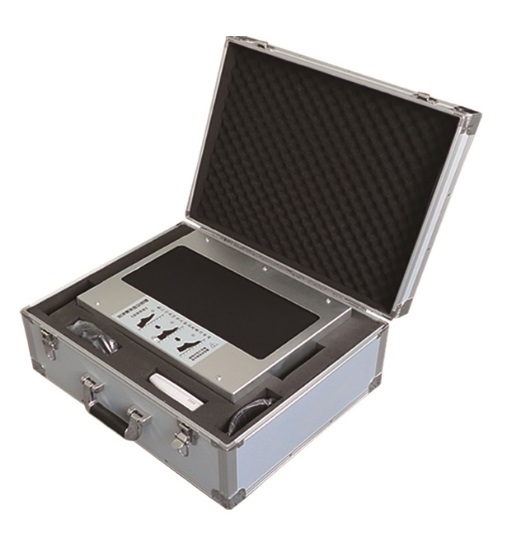}
	\caption{\normalfont{The machine (EverOS V2.0) used for shoeprints capturing. It captures the reflected scan after stepping on the strip. The acquisition system has been equipped with background noise removal.}}
	\label{fig:Foot_retrival_machine}
\end{figure*}
\begin{figure*}[ht!]
	\centering
	\includegraphics[width=18.5cm,height=8.1cm]{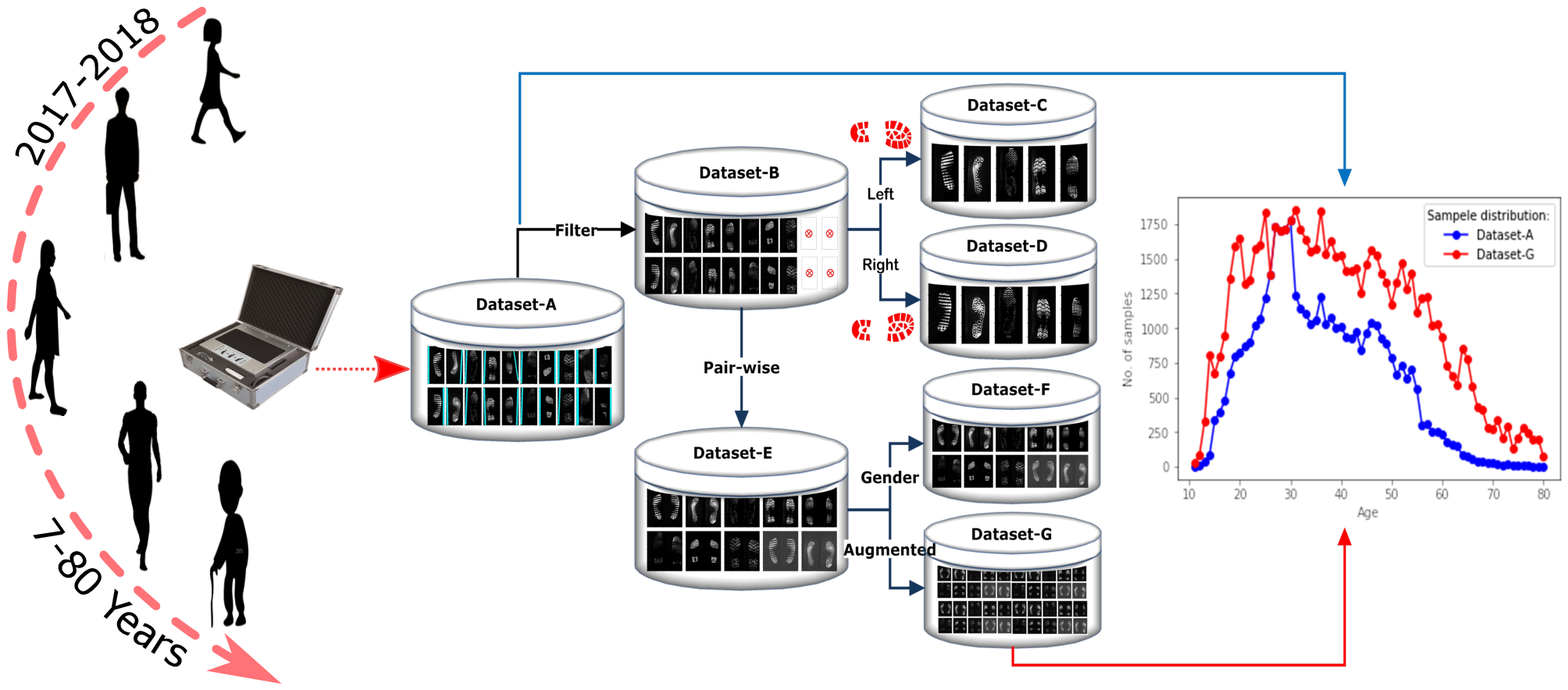}
	\caption{The overall generation of datasets generation and sample distributions, with brief descriptions, as which appear in Supplementary Table ~\ref{tab:datasets}. There are seven versions of shoeprints datasets including: Dataset-A, unprocessed and original shoeprints; Dataset-B, wherein the scale/ruler and poor-quality images are discarded manually; Dataset-C, which contains only left shoeprints; Dataset-D, which has only right shoeprints; Dataset-E, which is comprised of the horizontal concatenated left and right shoeprints into a single pair-wise shoeprint; Dataset-F, generated for gender-based classification and age estimation, and Dataset-G, the augmented dataset used to balance sample distribution in group-wise age prediction. The distributions of two datasets (Dataset-A and Dataset-G) are visualized.}
	\label{fig:Unprocessed_dataset}
\end{figure*}
\begin{figure*}[h!]
	\centering
	\includegraphics[height=3.1in,width=3.8in]{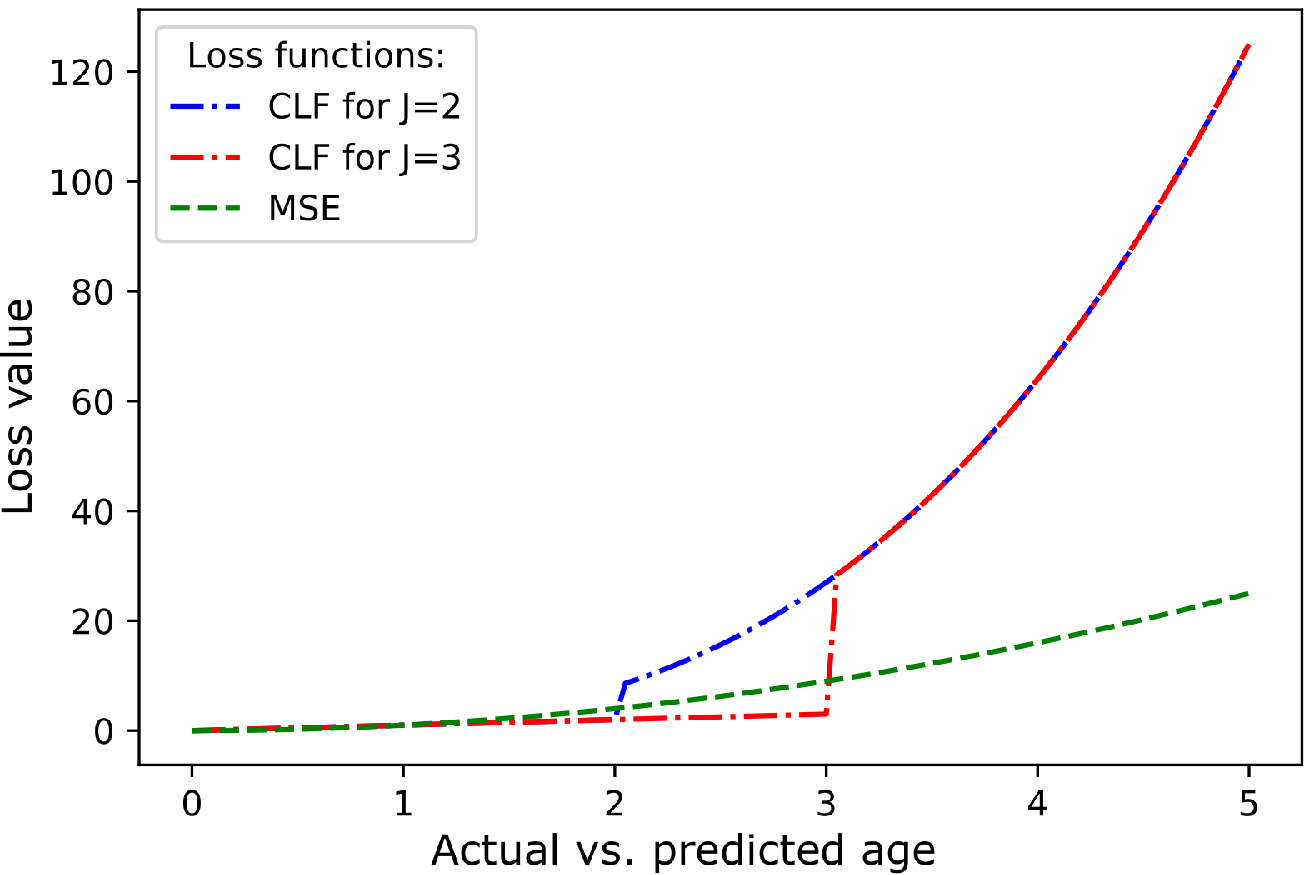}
	\caption{The customized loss function (CLF) versus prediction error in ShoeNet. This computation, unlike the linear regression, gives more weightage to the out-ranged values. Two ranges $(J \leq 2, J \leq 3)$ specified for CLF are shown as examples.}
	\label{fig:CLFvsMSE}
\end{figure*}
\begin{figure*}[h]
	\centering
	\includegraphics[height=2.8in,width=3in]{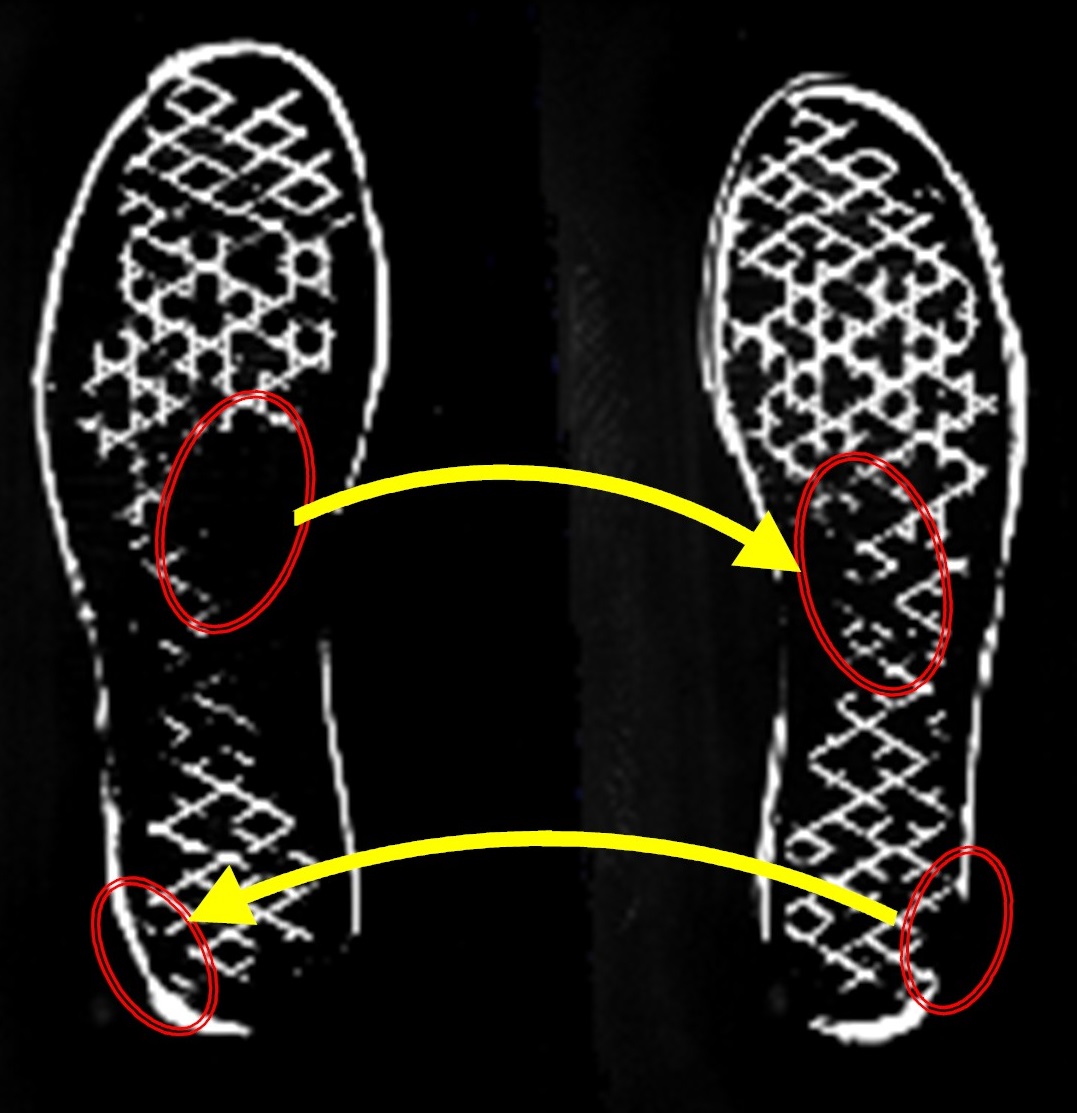}
	\caption{ShoeNet receives pair-wise shoeprints to capture age-related effects. The corresponding abrasion regions from paired shoeprints reflect gait and standing patterns, and variational effects to aging. By pair-wise left and right shoeprints, ShoeNet considers these regions of interest by convolving filters to capture the asymmetry for biological profile estimation.}
	\label{fig:correspong_scratching}
\end{figure*}
\begin{figure*}[!]
	\centering
	\includegraphics[width=16cm,height=12cm]{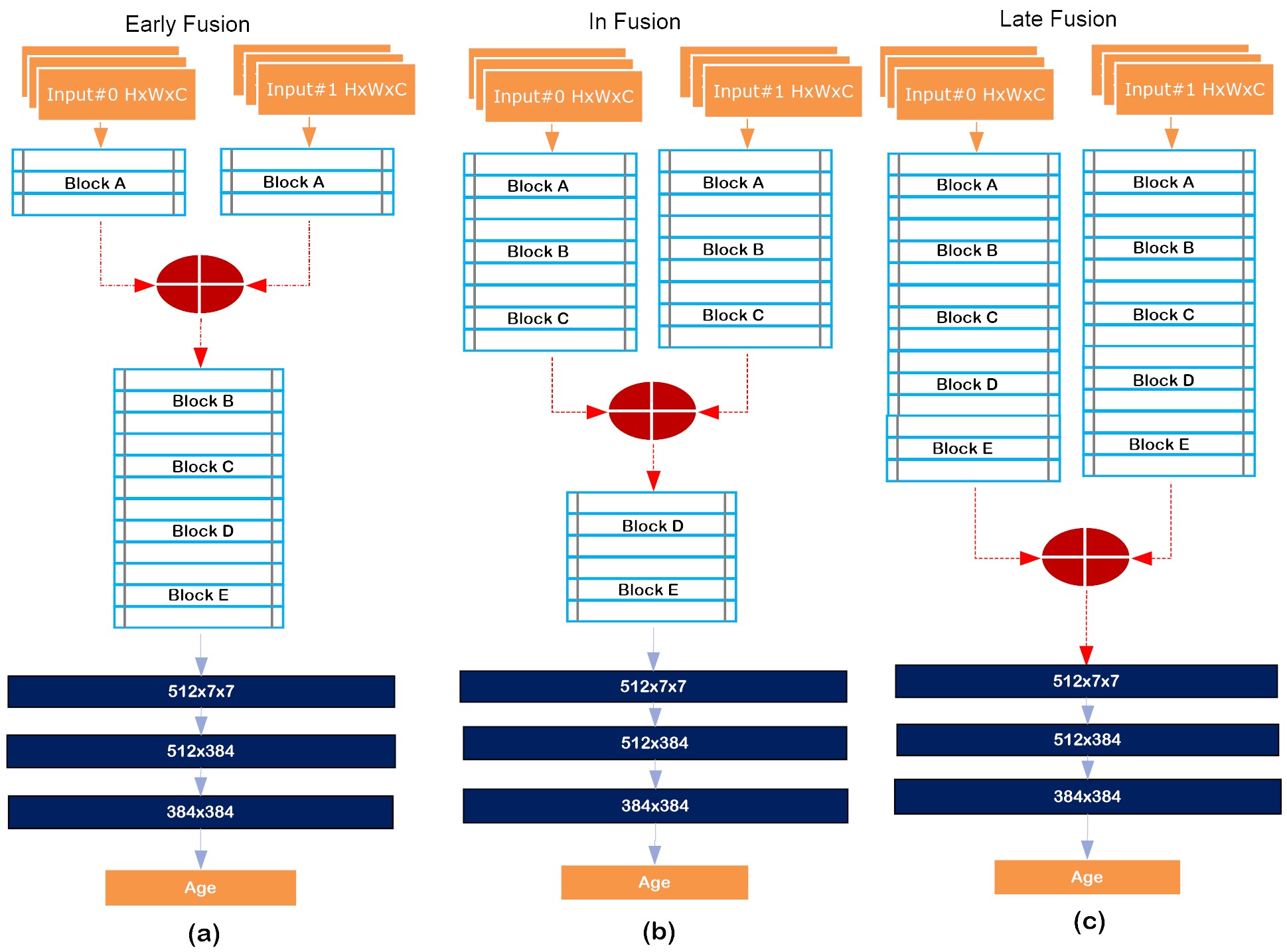}
	\caption{The architecture of fusion-models. Each model receives two input images and fuses at different levels. All the sub-networks have similar fully connected layers and a single output neuron for age prediction. (a) Early-fusion fuses the two input images just after the first layer of the convolution operations. Prior to fusion in Block-A, the network extracts features from the left and right shoeprints separately and then combines the features. In the early fusion, the features from the parallel networks are concatenated into a stack of features for further convolutions from Block-B to Block-E. (b) In-fusion merges the feature map in the middle after Block-C to utilize both isolated and joint representations, followed by convolutions of Block-E and Block-E. (c) Late-fusion merges the twin networks just after Block-E, followed by fully connected layers.}
	\label{fig:Fusion Model}
\end{figure*}
\begin{figure*}[h!]
	\centering
	\includegraphics[ width=3.4in]{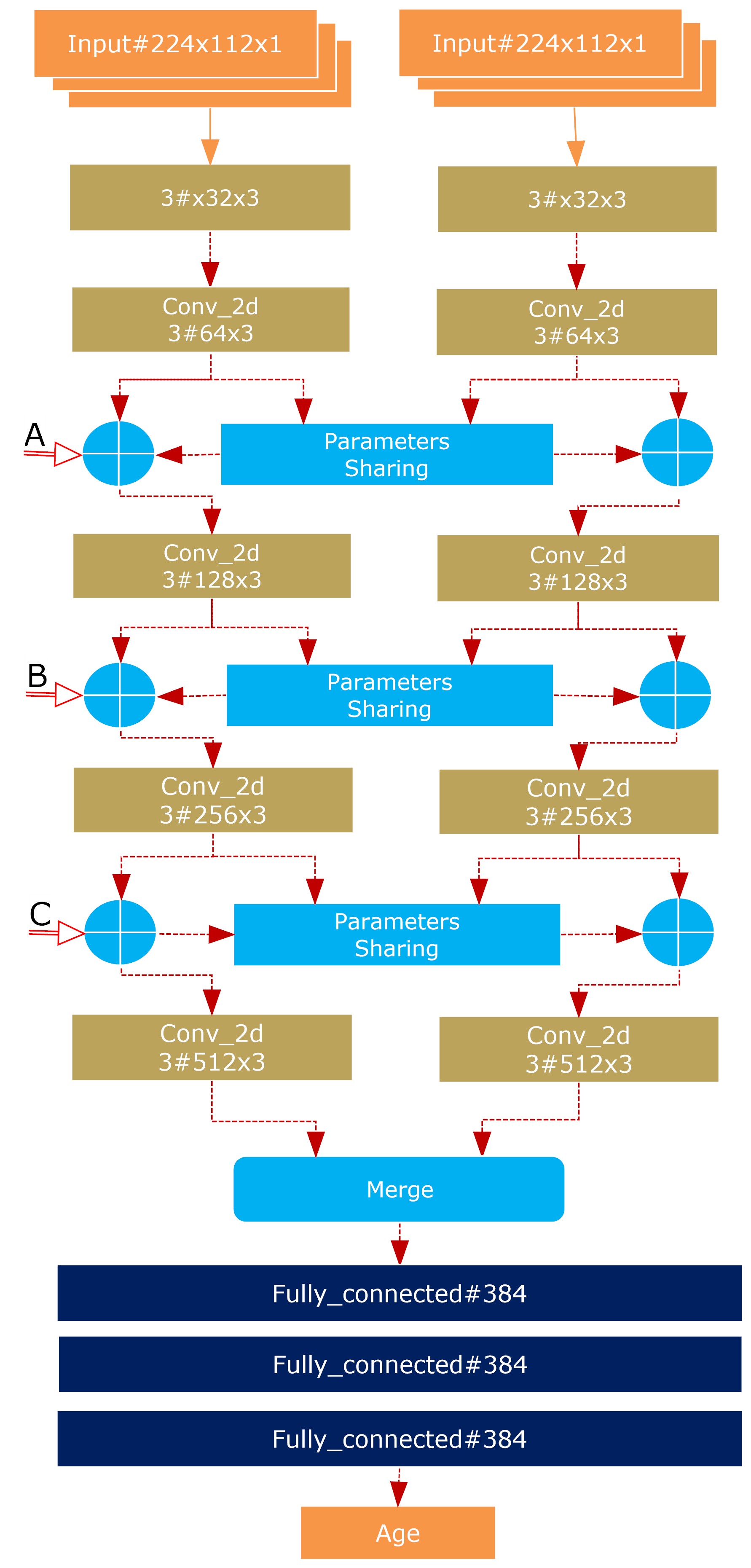}
	\caption{The architecture of multi-modal-CNN (MM-CNN) is demonstrated. The network accepts two separate images (left and right shoeprints) at one time. The same structures and dimensions are applied for both input images in two parallel channels. MM-CNN shares information at different levels of the network including early level (level-A), mid-level (level-B) and late-level (level-C). The information sharing at different levels enables parallel networks to exchange features of significance regarding age estimation. In all the cases, the parallel networks concatenated into a single network, which further passes through fully connected layers and ends with a single neuron for age prediction.}
	\label{fig:multi-model}
\end{figure*}
\begin{figure*}[h!]
	\centering
	\subfigure[Left shoeprint pressure distributions for eight divided regions]
	{\raisebox{16mm}{\includegraphics[height=1.4in,width=.7in]{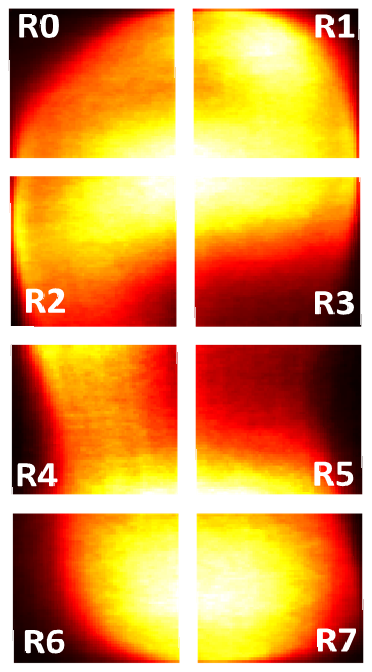}}
		\includegraphics[height=2.8in,width=2.35in]{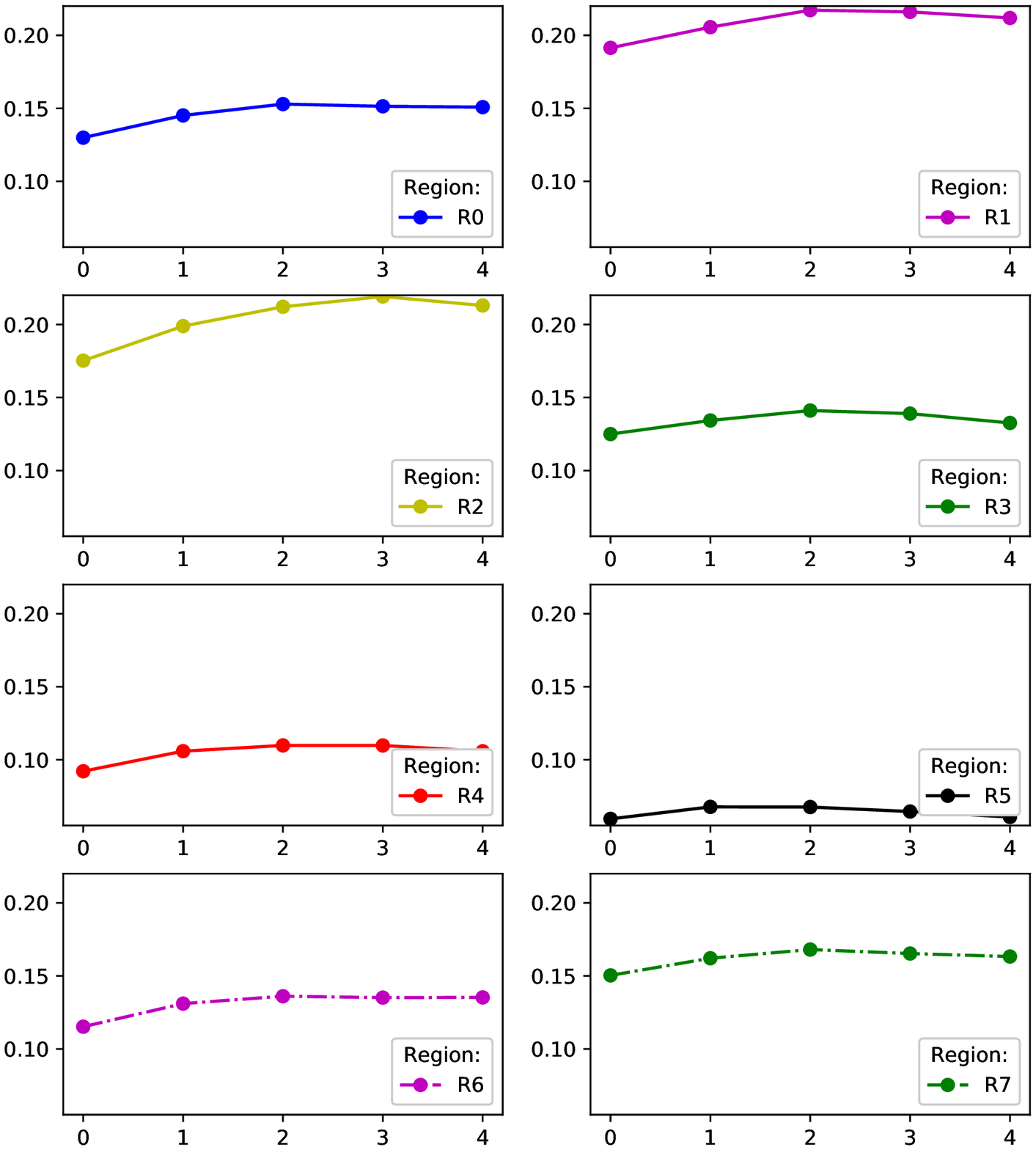}
		\label{fig:left shoeprint-same-scale}
	}
	\subfigure[Right shoeprint pressure distributions for eight divided regions]{
		\includegraphics[height=2.8in,width=2.35in]{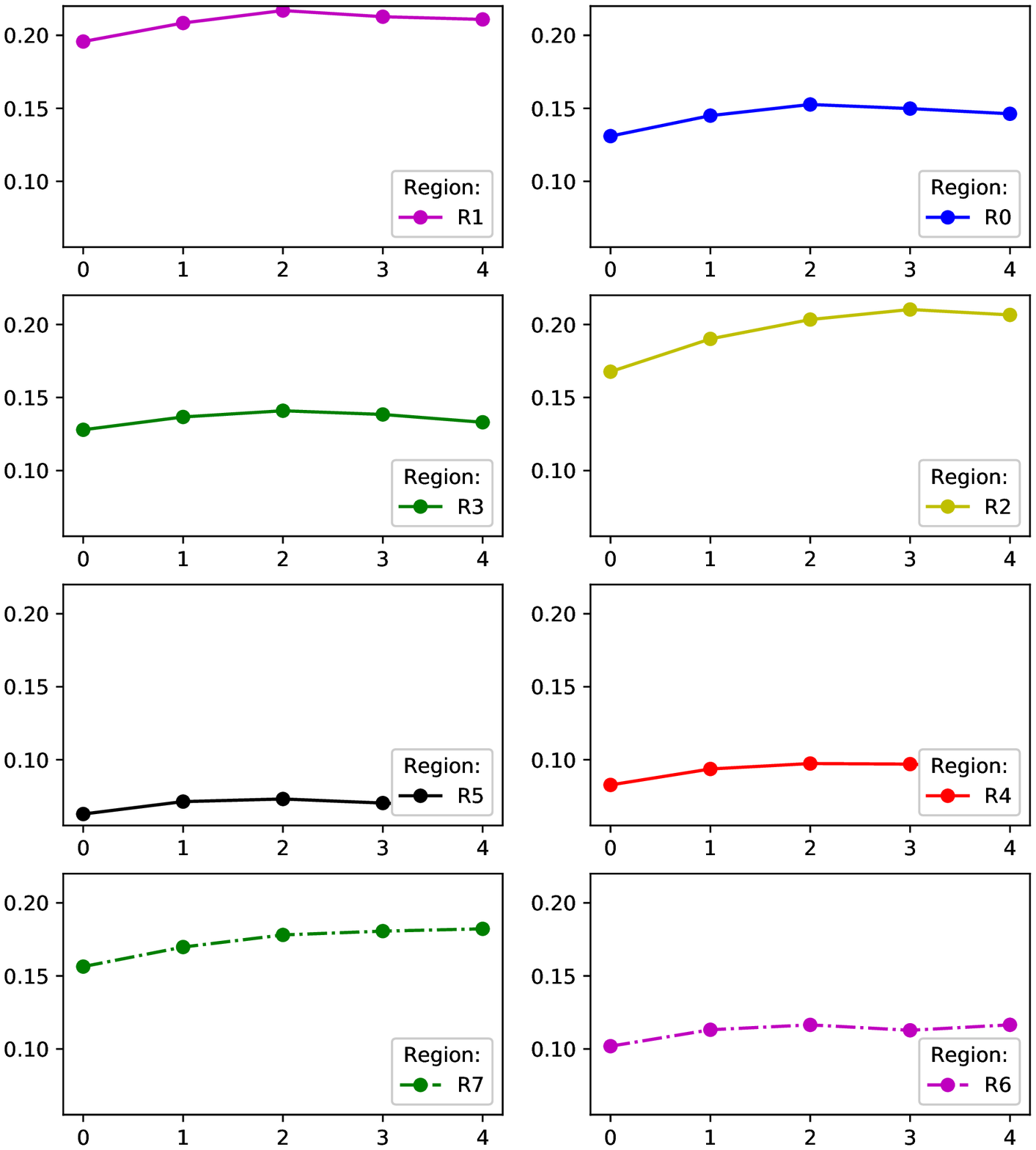}
		\raisebox{18mm}{\includegraphics[height=1.4in,width=.7in]{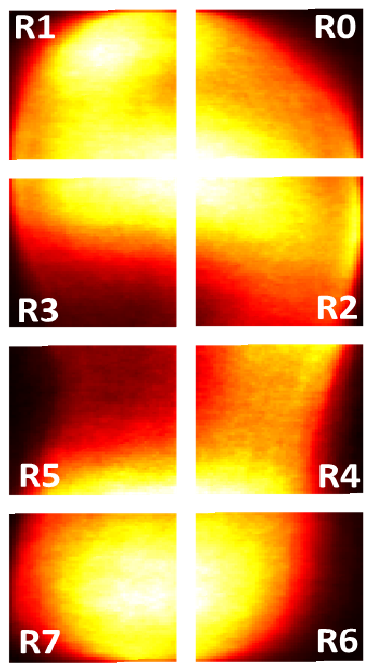}}
		\label{fig:right shoeprint-same-scale}
	}
	\caption{Category-wise pressure distribution and variations versus aging. All the regions are scaled into the same upper and lower boundaries and divided into eight regions for (a) left shoeprints and (a) right shoeprints.}
	\label{fig:eight-regions-curves-left-right-all-same-scale}
\end{figure*}
\begin{figure*}[!]
	\centering
	\includegraphics[width=18cm,height=9cm]{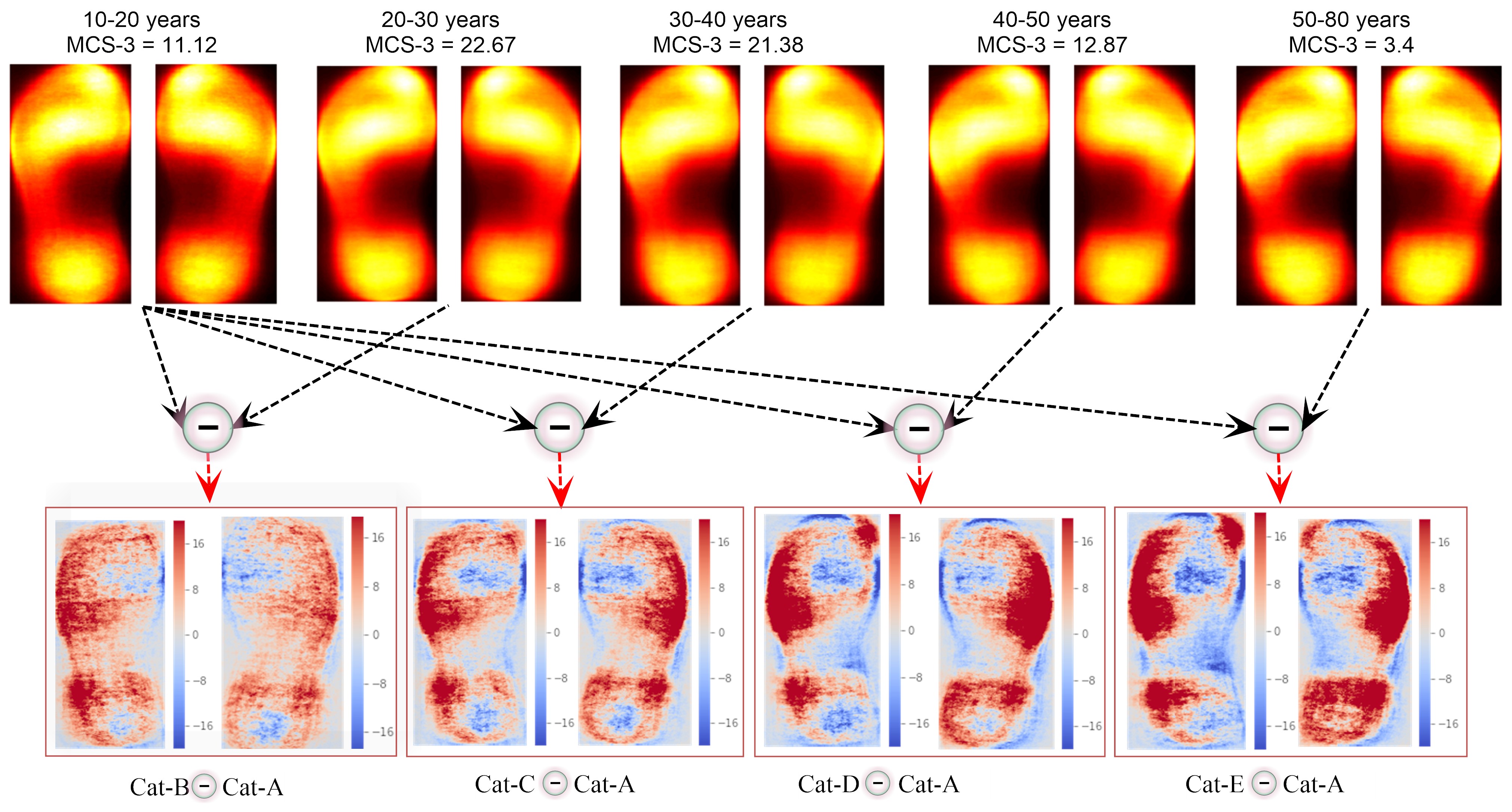}
	\caption{Pressure differences between the subjects below 20 years and the rest of the categories subjects.}
	\label{fig:subtraction_of_early_category_from_the_rest_of_cat}
\end{figure*}
\begin{figure*}[!]
	\centering
	\includegraphics[width=18cm,height=9cm]{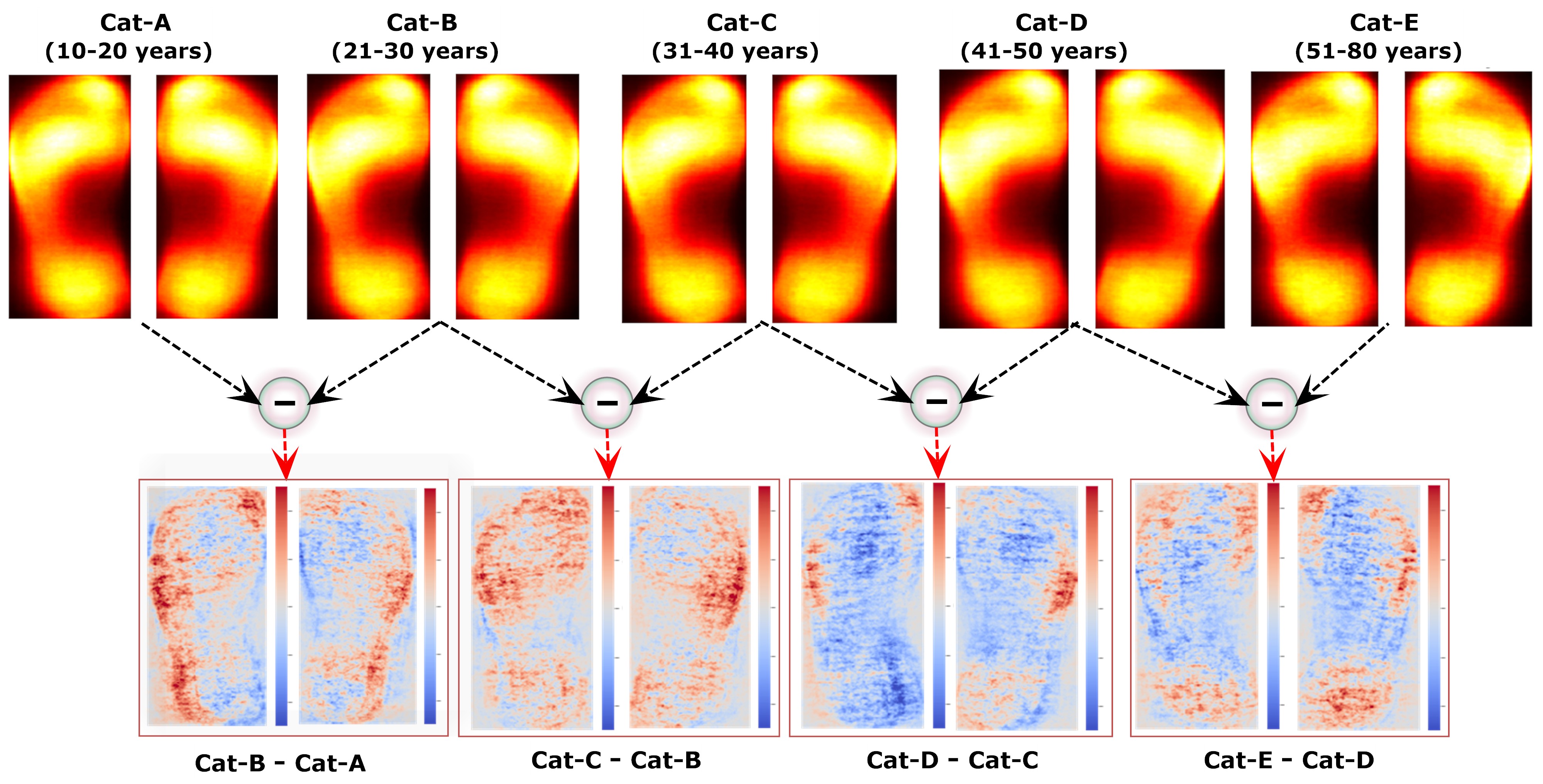}
	\caption{The shoeprint pressure distribution versus age in males. The first row has superimposed male shoeprints of five categories (category-A, B, C, D, E) for age ranges 10-20, 21-30, 31-40, 41-50, and 51-80 years, respectively. The second row depicts the lower category subtraction from the upper category; for instance, Category-B minus Category-A.}
	\label{fig:male groupwise pressure distribution subtraction}
\end{figure*}
\begin{figure*}[!]
	\centering
	\includegraphics[width=18cm,height=9cm]{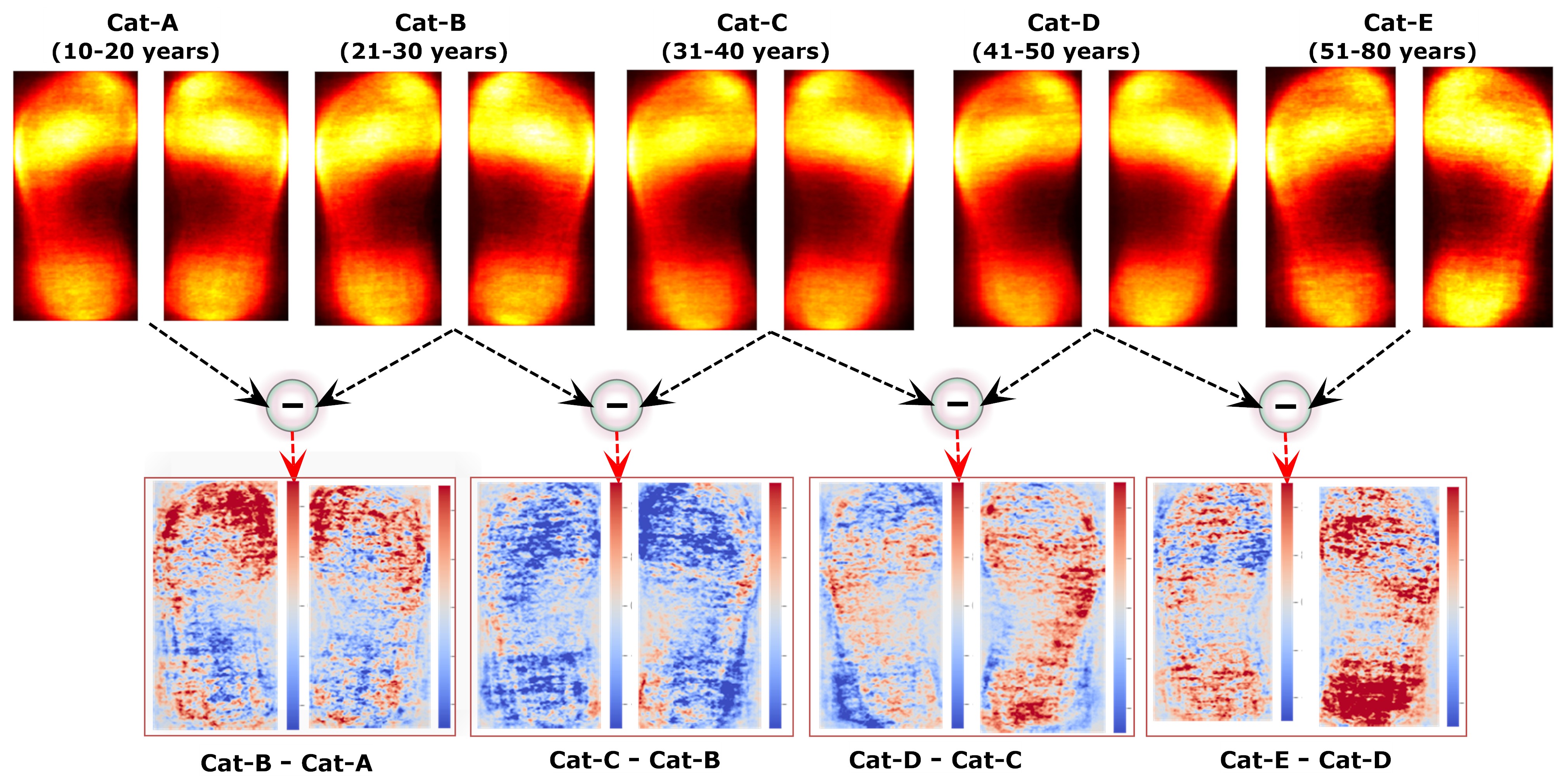}
	\caption{The shoeprint pressure distribution versus age in females. The first row has superimposed male shoeprints of five categories (category-A, B, C, D, E) for age ranges 10-20, 21-30, 31-40, 41-50, and 51-80 years, respectively. The second row depicts the lower category subtraction from the upper category.}
	\label{fig:female groupwise pressure distribution subtraction}
\end{figure*}
\begin{figure*}[h!]
	\centering
	\subfigure[Same scale for MnF left shoeprints]{
		\includegraphics[height=2.9in,width=2.9in]{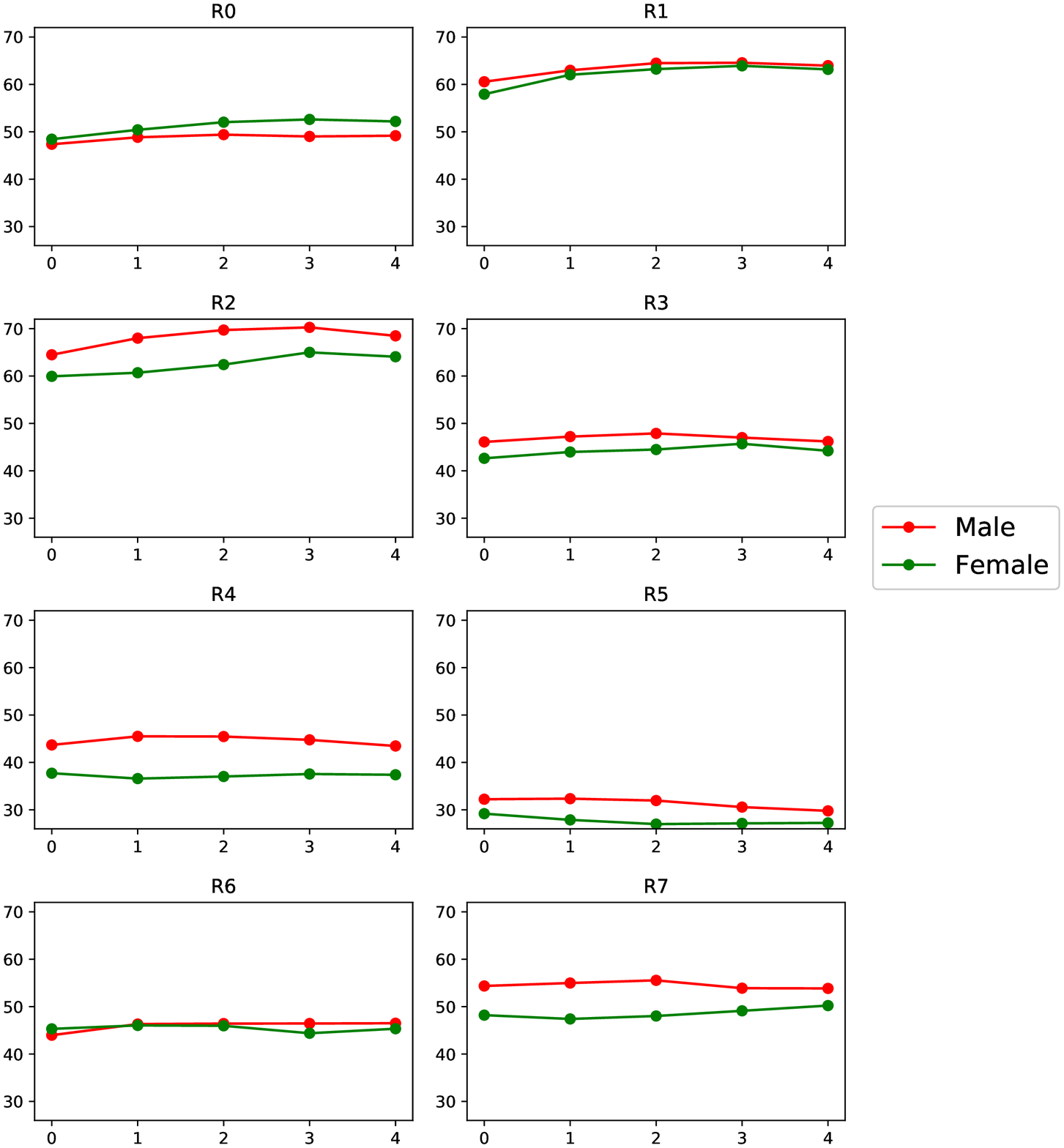}
		\label{fig:male joint female left_same_scale}
	}
	\hskip -2ex
	\subfigure[Same scale for MnF right shoeprints]{
		\includegraphics[height=2.9in,width=2.6in]{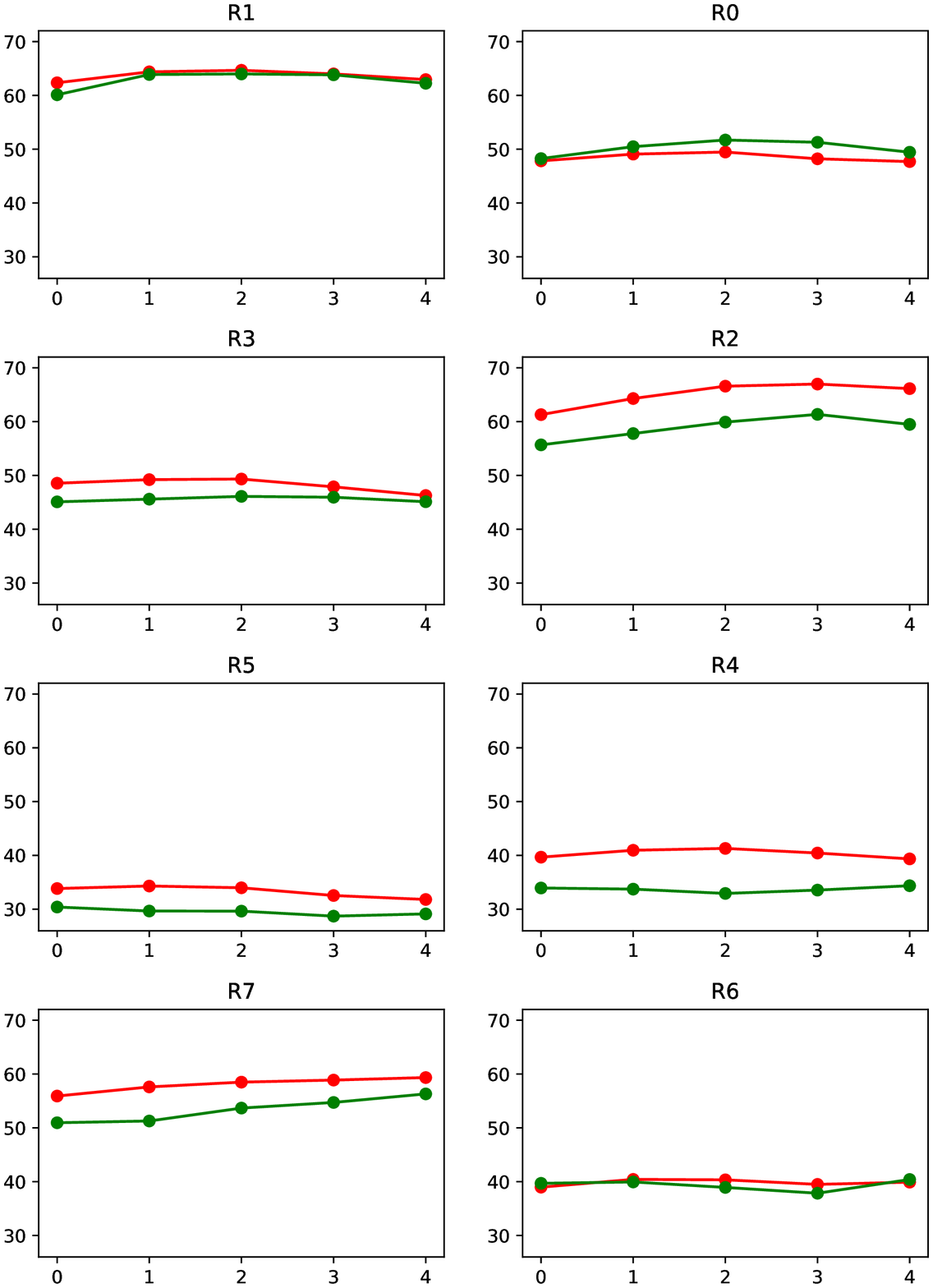}
		\label{fig:male joint female right_same_scale}
	}
	
	\caption{Region-wise and gender-wise pressure distributions and variations with the same scale, and the same upper and lower limits, versus age for the (a) left shoeprints and the (b) right shoeprints.}
	\label{fig:eight-regions-curves-gender-left-right-same-scale}
\end{figure*}
\begin{figure*}[h!]
	\centering
	\subfigure[]{
		\includegraphics[height=2.5in,width=3.1in]{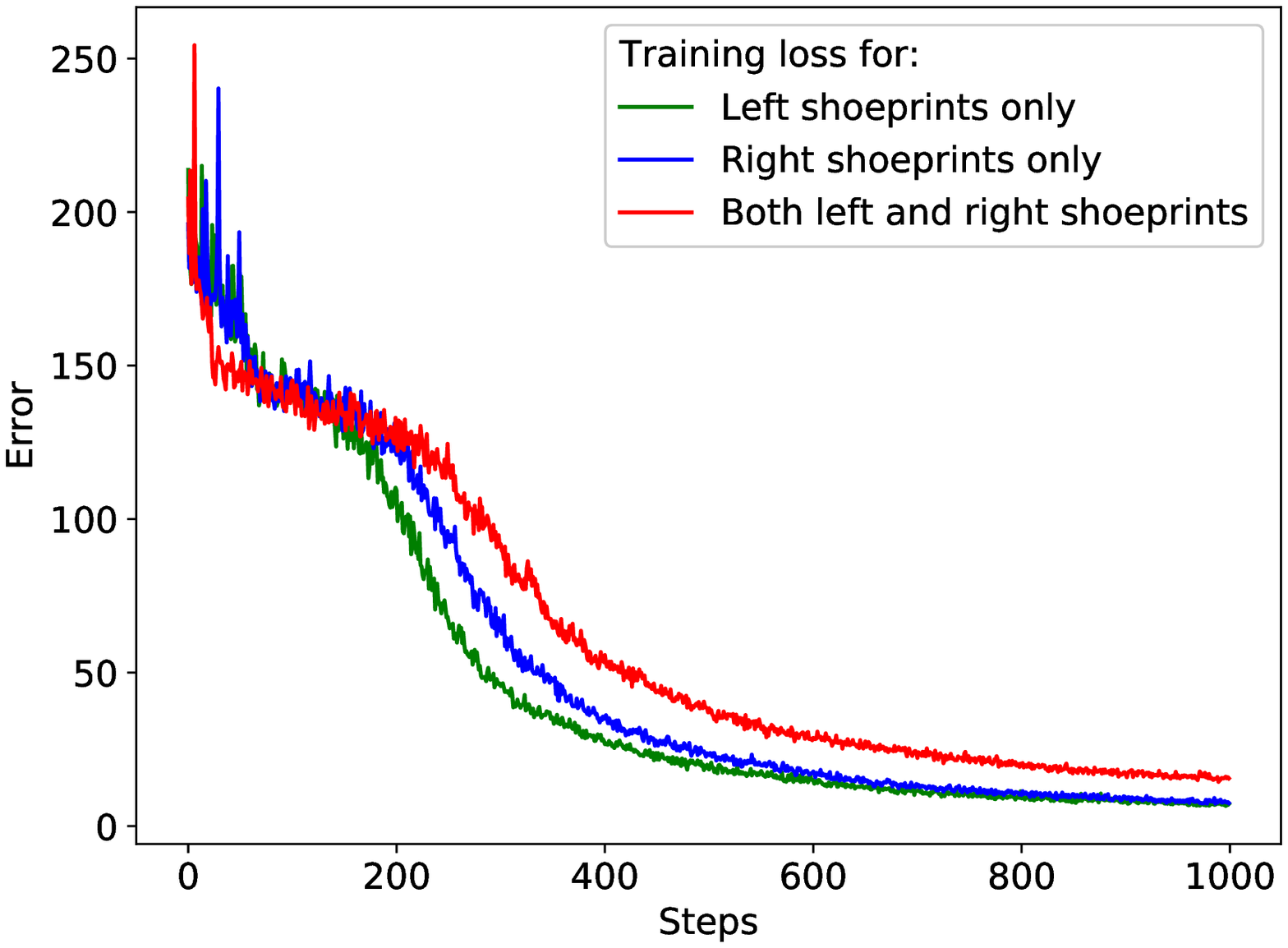}
		\label{fig:left-to-right-training}
	}
	\subfigure[]{
		\includegraphics[height=2.5in,width=3.1in]{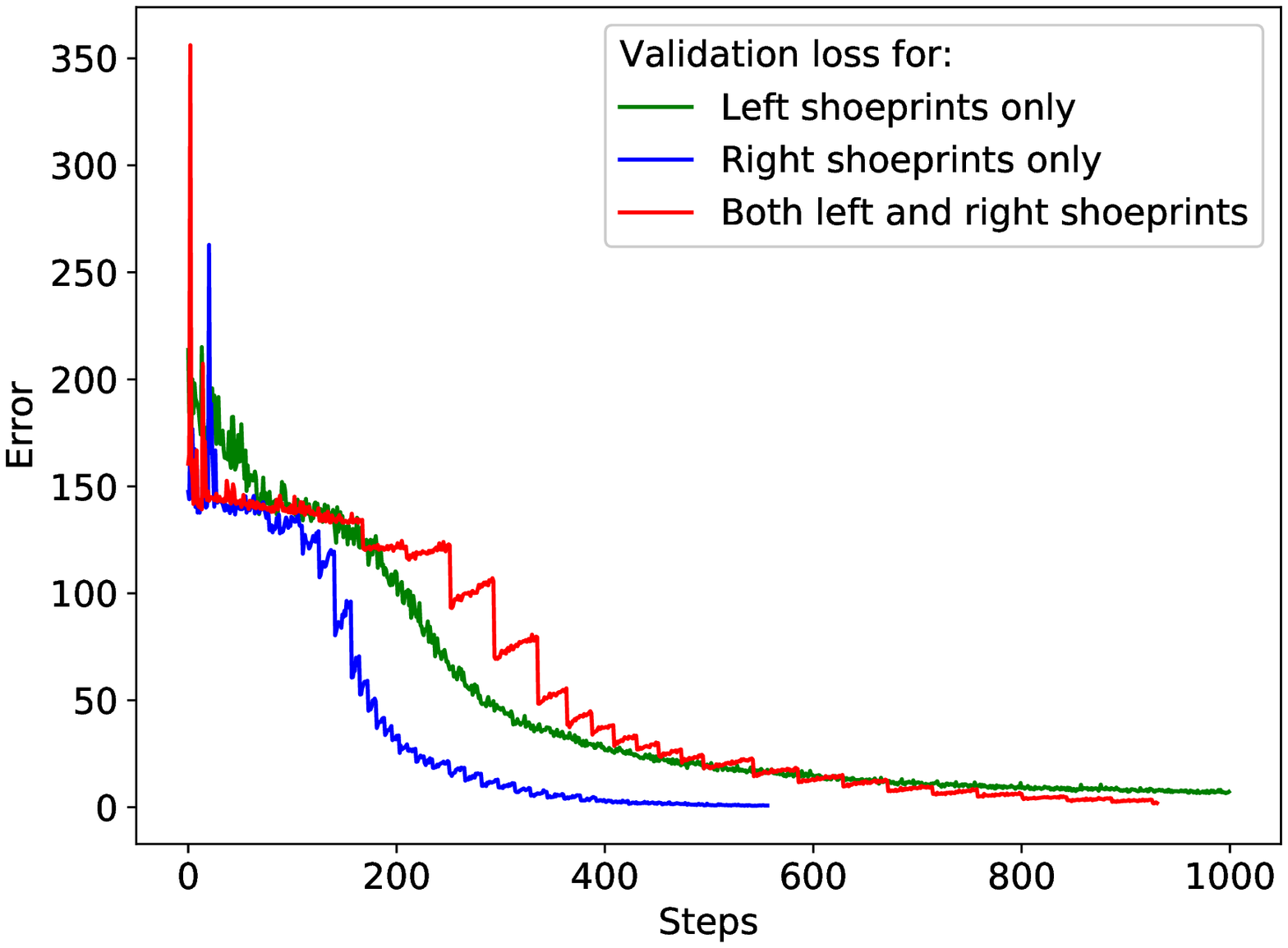}
		\label{fig:left-to-right-validation}
	}
	\caption{(a) The learning curves of the three (left, right, left-to-right) networks in LR-CNN. The network trained with mingled L\&R shoeprints has a steady but slow drop in the error rate while the networks based on left or right shoeprints have swift drops in the error rate. Such variations also affect the performance during testing (Table ~\ref{tab:tab2}), where the network trained solely on left or right shoeprint has a higher score than the network trained on both shoeprints. (b) Similarly, the corresponding validation errors are shown for the three trained network modalities, which demonstrates similar trends as a slow decline in error for mingled L\&R shoeprints.}
	\label{fig: LR-CNN}
\end{figure*}
\begin{figure*}[h!]
	\centering
	\subfigure[]{
		\includegraphics[height=2.5in,width=3.1in]{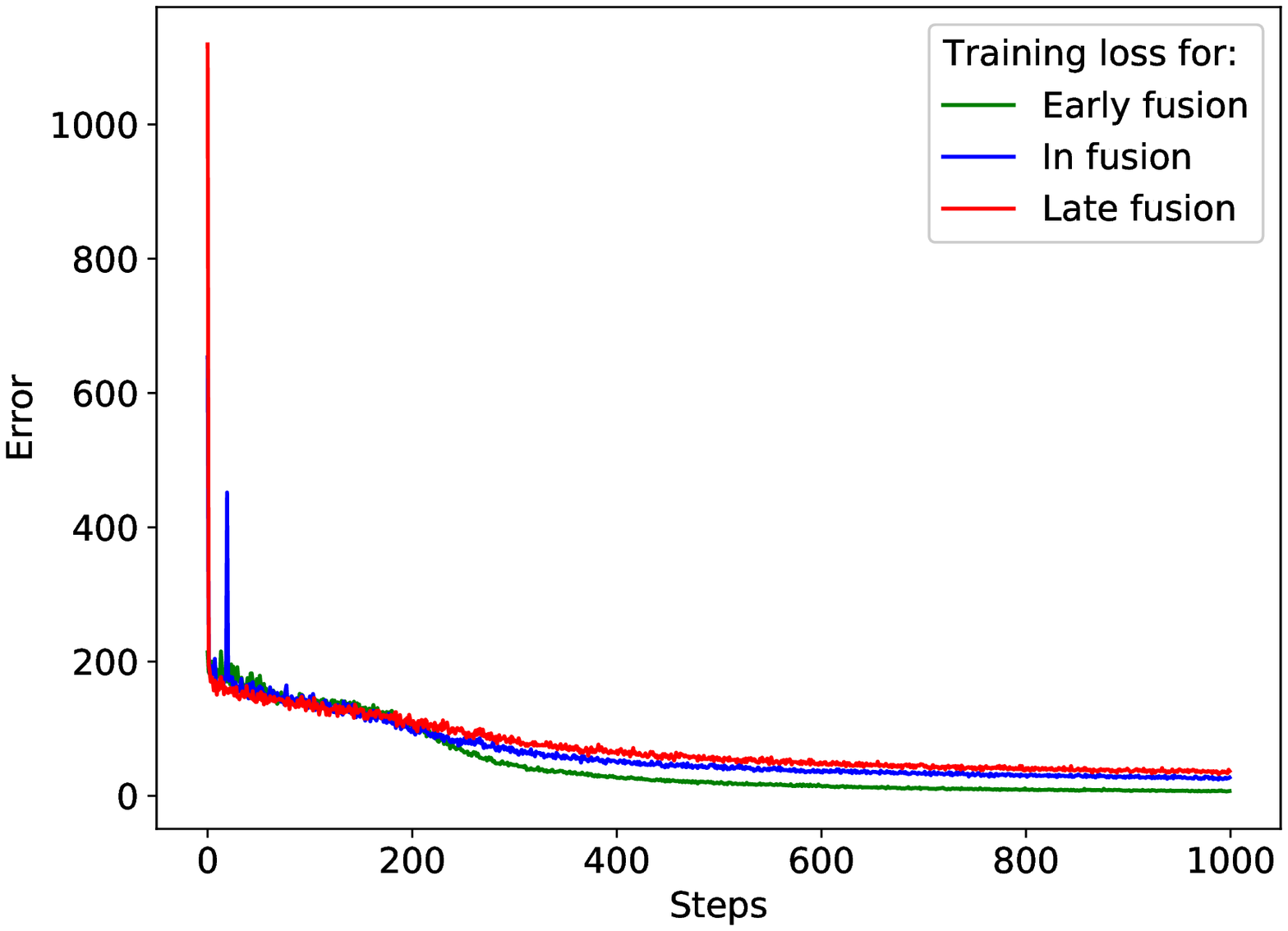}
	}
	\subfigure[]{
		\includegraphics[height=2.5in,width=3.1in]{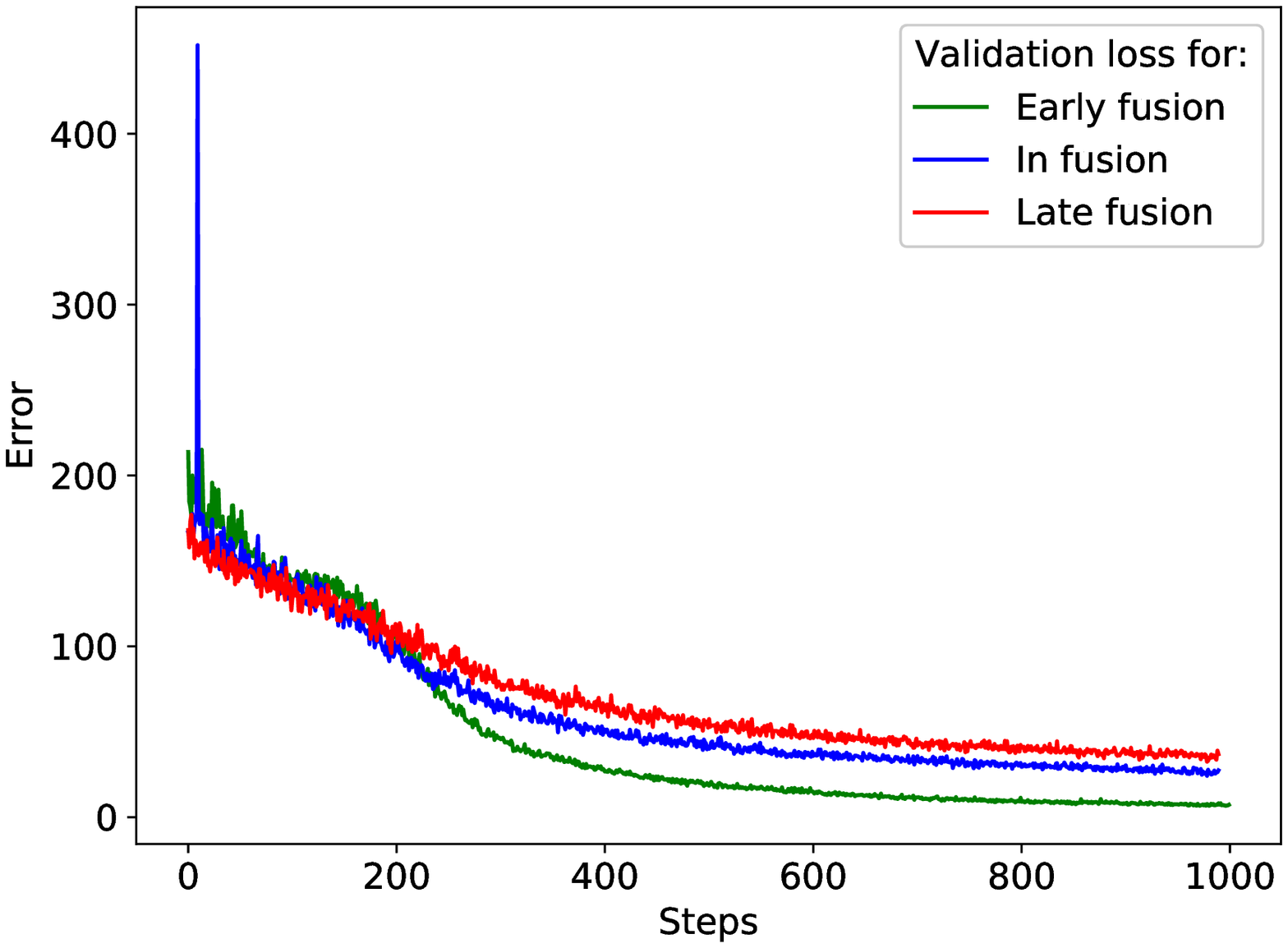}
	}
	\caption{The learning curves of fusion modalities for (a) training and (b) validation. Early fusion and in-fusion show slightly better results in both training and validation.}
	\label{fig:fusion-left-to-right_training_validation}
\end{figure*}
\begin{figure*}[h!]
	\centering
	\subfigure[]{
		\includegraphics[height=2.5in,width=3.1in]{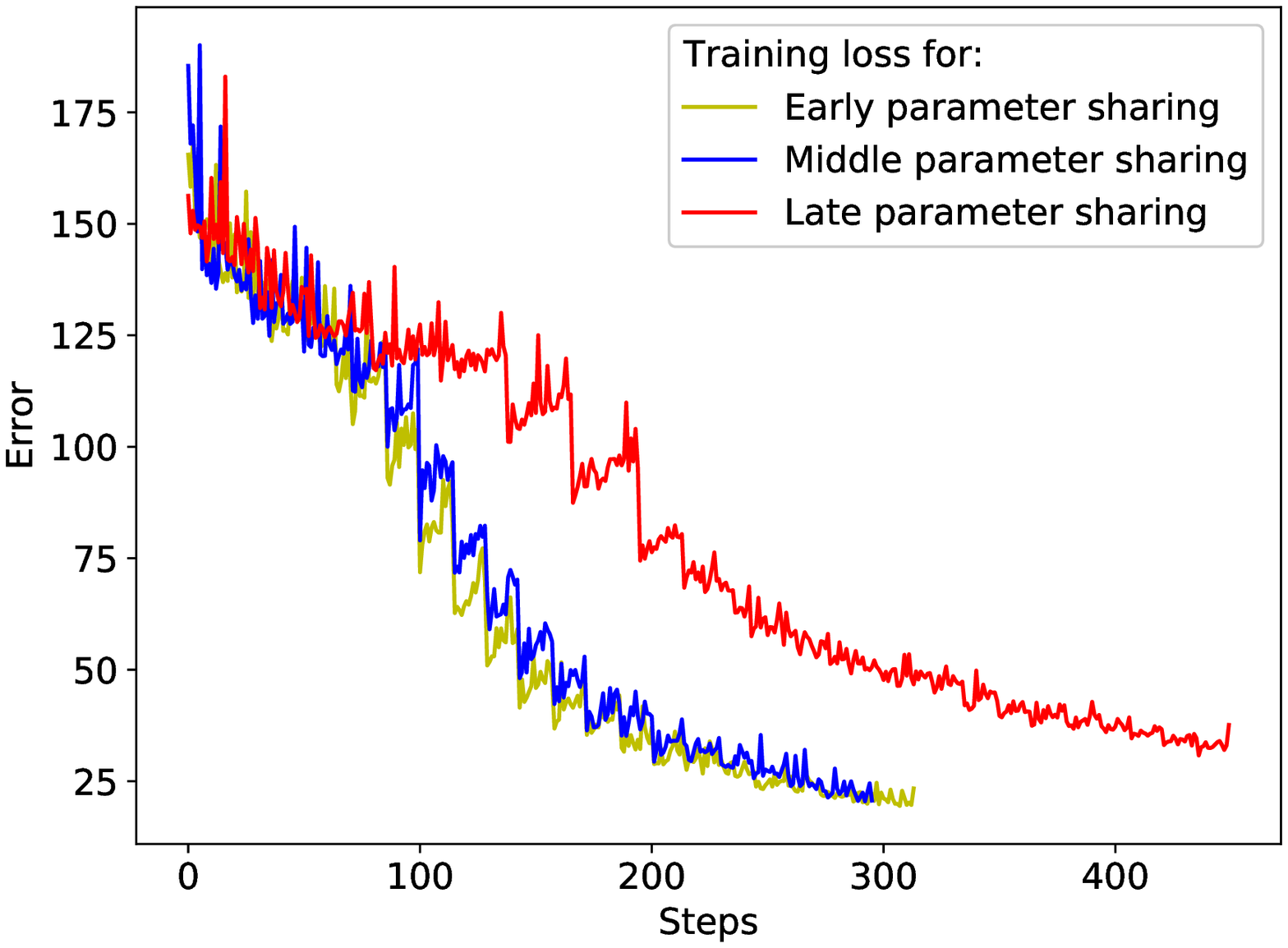}
		
	}
	\subfigure[]{
		\includegraphics[height=2.5in,width=3.1in]{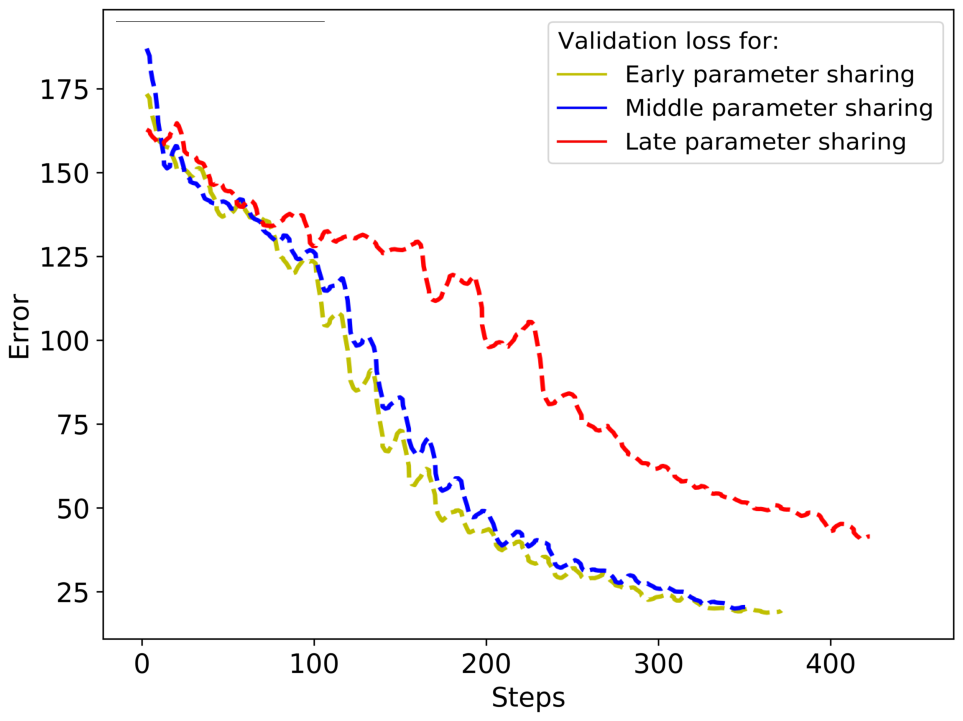}
	}
	\caption{The learning curves for (a) training and (b) validation of early parameter sharing (Early-PS), middle parameter sharing (Mid-PS) and late parameter sharing (Late-PS). The early sharing network rapidly converges toward zero, while the late parameter sharing has a slow decrease in error rate caused by complex twin structural networks. During the validation, the late parameter sharing has an even more distinguished slow tendency in error decrease.}
	\label{fig:multi-model-sharing-training-and-validation}
\end{figure*} 

\end{document}